\documentclass[final,10pt]{elsarticle}

\usepackage{graphicx}
\usepackage{amssymb}
\usepackage{amsmath}
\usepackage{booktabs}
\usepackage{multirow}
\usepackage{caption}
\usepackage{xspace}
\usepackage[table]{xcolor}
\usepackage{float}
\usepackage{colortbl}
\usepackage{makecell}
\usepackage{listings}
\usepackage{algorithm}
\usepackage{algorithmic}
\usepackage{subcaption}
\usepackage{stfloats}
\usepackage{rotating}

\lstdefinestyle{shell}{
  basicstyle=\ttfamily\tiny,
  backgroundcolor=\color{gray!10},
  frame=single,
  breaklines=true,
  postbreak=\mbox{\textcolor{red}{$\hookrightarrow$}\space},
  showstringspaces=false
}

\makeatletter
\def\ps@pprintTitle{%
  \let\@oddhead\@empty
  \let\@evenhead\@empty
  \def\@oddfoot{%
    \parbox{\textwidth}{\footnotesize
      Accepted manuscript for publication in Pattern Recognition.\\
      DOI: https://doi.org/10.1016/j.patcog.2026.113824 \\
      © 2026 The Authors. Published by Elsevier Ltd. This is an open access article distributed under the terms of the Creative Commons CC-BY license 4.0. https://creativecommons.org/licenses/by/4.0/
    }%
  }%
  \let\@evenfoot\@oddfoot
}
\makeatother

\journal{Pattern Recognition}

\makeatletter
\DeclareRobustCommand\onedot{\futurelet\@let@token\@onedot}
\def\@onedot{\ifx\@let@token.\else.\null\fi\xspace}

\def\ie{\emph{i.e}\onedot}

\def\etal{\emph{et al}\onedot}

\AtEndPreamble{
    \usepackage[capitalize]{cleveref}
    \crefname{section}{Sec.}{Secs.}
    \Crefname{section}{Section}{Sections}
    \Crefname{table}{Table}{Tables}
    \crefname{table}{Tab.}{Tabs.}
}

\begin{document}

\begin{frontmatter}

\title{BIR-Adapter: A parameter-efficient diffusion adapter for blind image restoration}

\author[org1,org3]{{\small Cem Eteke}\corref{cor1}}
\ead{cem.eteke@tum.de}
\author[org2,org3]{{\small Alexander Griessel}}
\author[org2,org3]{{\small Wolfgang Kellerer}}
\author[org1,org3]{{\small Eckehard Steinbach}}

\affiliation[org1]{
    organization={Chair of Media Technology, Munich Institute of Robotics and Machine Intelligence}
}
\affiliation[org2]{
    organization={Chair of Communication Networks}
}

\affiliation[org3]{
    organization={School of Computation, Information and Technology, Technical University of Munich},
    city={Munich},
    postcode={80333}, 
    country={Germany}
}

\cortext[cor1]{Corresponding author}


\begin{abstract}
We introduce the BIR-Adapter, a parameter-efficient diffusion adapter for blind image restoration. Diffusion-based restoration methods have demonstrated promising performance in addressing this fundamental problem in computer vision, typically relying on auxiliary feature extractors or extensive fine-tuning of pre-trained models. Building on the observation that large-scale pretrained diffusion models can retain informative representations under image degradations, BIR-Adapter introduces a parameter-efficient, plug-and-play attention mechanism that substantially reduces the number of trained parameters. To further improve reliability, we adapt a sampling guidance mechanism that mitigates hallucinations during restoration. Experiments on synthetic and real-world degradations demonstrate that BIR-Adapter achieves competitive, and in several settings superior, performance compared to state-of-the-art methods while requiring up to $36 \times$ fewer trained parameters. Moreover, the adapter-based design enables integration into existing models. We validate this generality by extending a super-resolution–only diffusion model to handle additional unknown degradations, highlighting the adaptability of our approach for broader image restoration tasks.
\end{abstract}

\begin{graphicalabstract}
    \centering
    \begin{minipage}[t]{0.518\textwidth}
      \vspace{1pt}
      \centering
      \includegraphics[width=\linewidth,height=0.48\textheight,keepaspectratio]{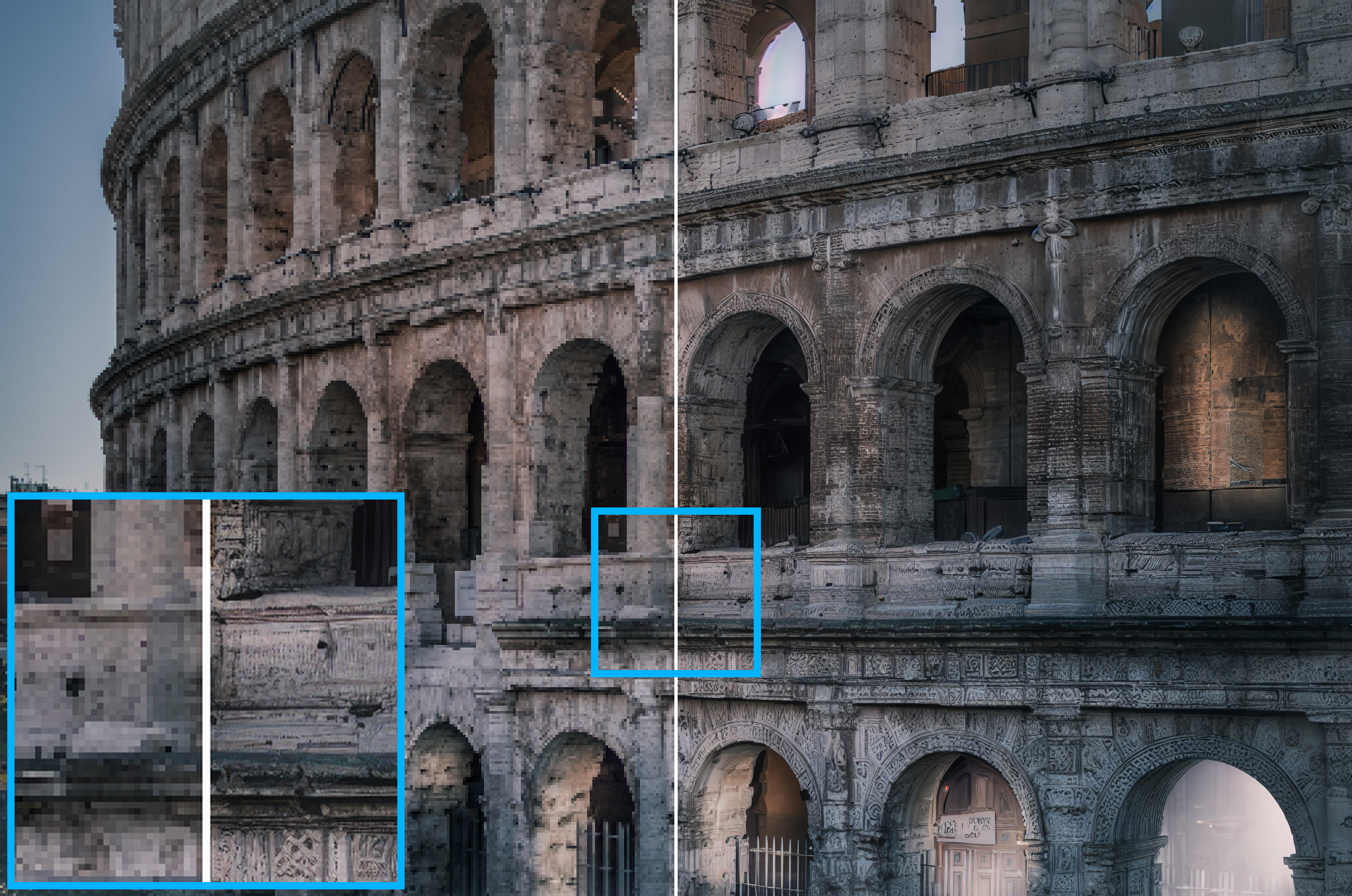}\\[3pt]
      \includegraphics[width=\linewidth,height=0.48\textheight,keepaspectratio]{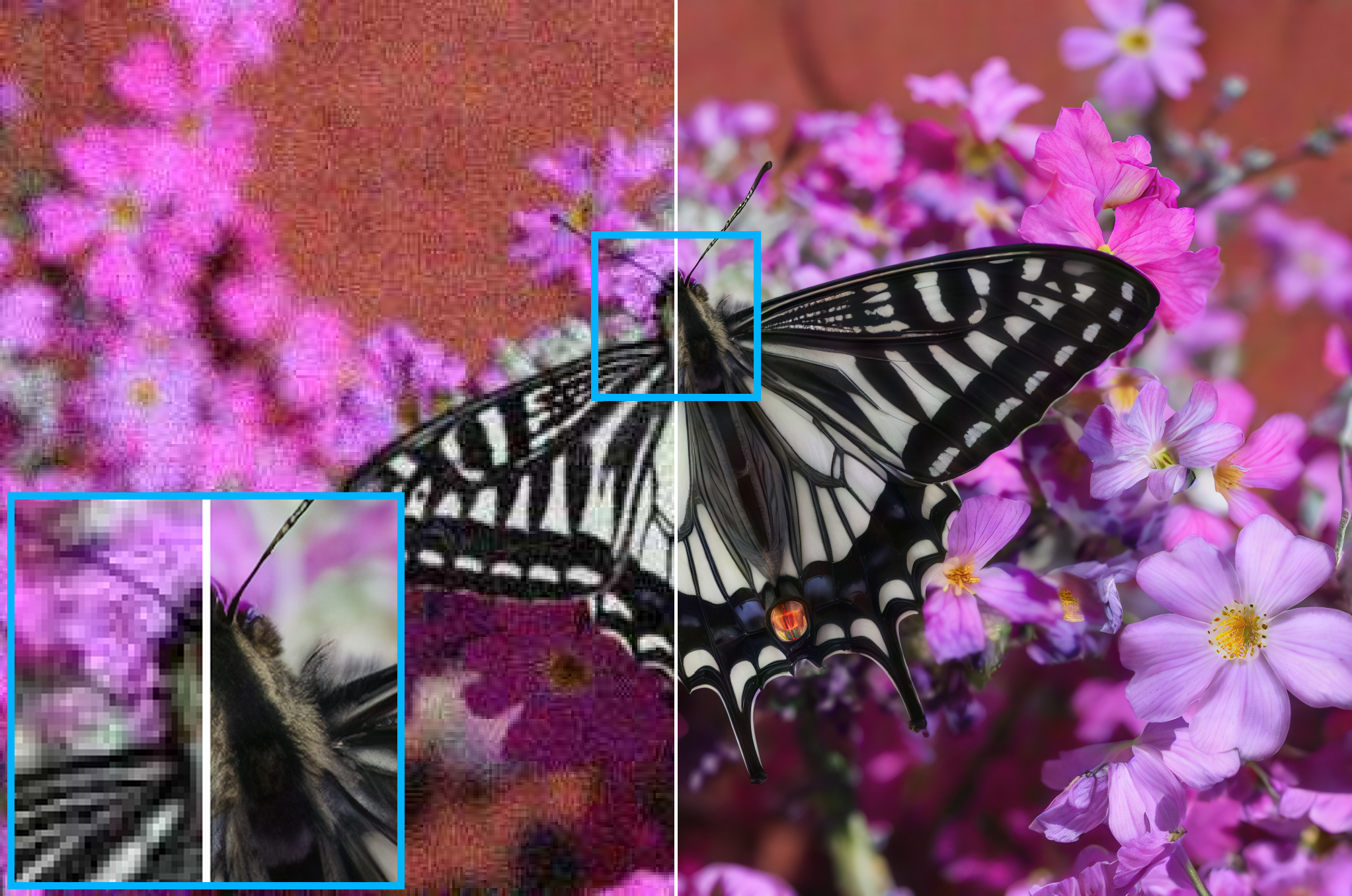}
    \end{minipage}%
    \hspace{1pt}
    \begin{minipage}[t]{0.465\textwidth}
      \vspace{1pt}
      \centering
      \includegraphics[width=\linewidth,height=0.97\textheight,keepaspectratio]{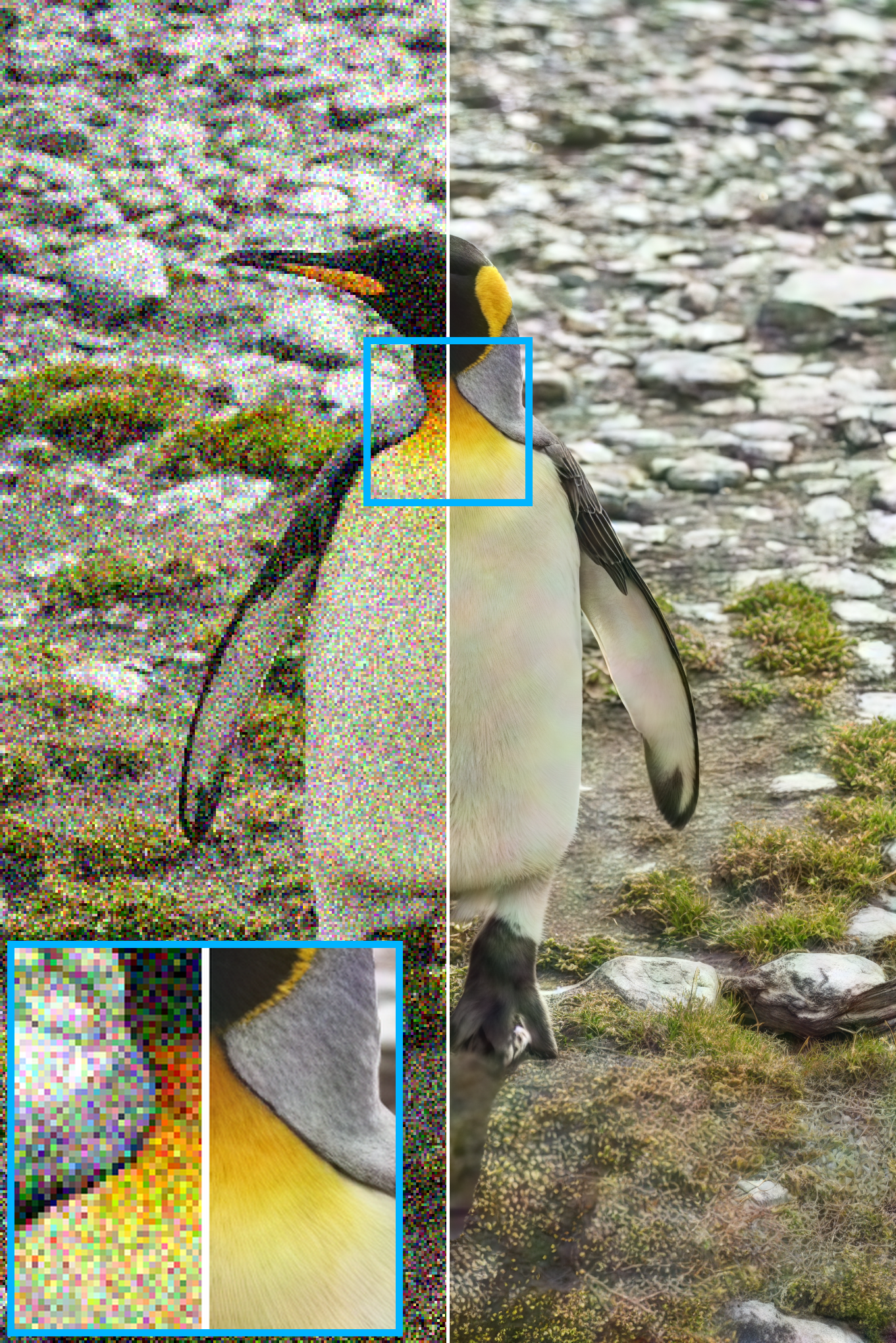}
    \end{minipage}
    \captionsetup{hypcap=false}
    \captionof{figure}{Example restoration results of BIR-Adapter on images degraded with $4\times$ downsampling. The lower-left image is further degraded with blur, white noise, and JPEG compression, while the right-hand image includes additional white noise.}
    \label{fig:hero}
    \vspace{2ex}
\end{graphicalabstract}

\begin{highlights}
\item Provides empirical evidence that diffusion models retain aligned representations under unknown image degradations.
\item Leverages internal features to enable blind image restoration without auxiliary feature extractors.
\item Introduces a parameter-efficient restoring attention mechanism that couples degraded and denoised latent features.
\item Achieves strong performance with significantly fewer additional trained parameters.
\item Preserves a plug-and-play adapter design that generalizes across diffusion-based restoration methods.
\end{highlights}


\end{frontmatter}


\section{Introduction}
\label{sec:intro}

Blind image restoration is the process of restoring the clean version of a degraded observation where the underlying degradation process is unknown. This problem is fundamentally challenging due to the wide range of degradations, including noise, blur, downsampling, compression artifacts, and complex cascades thereof. Supervised deep neural networks demonstrated promising performance by learning a mapping between degraded and clean images~\cite{zhai2023comprehensive}. Furthermore, with the advent of generative methods such as generative adversarial networks (GAN)~\cite{goodfellow2020generative} and diffusion models~\cite{dhariwal2021diffusion}, the restoration performance has been further improved as generative modeling enables synthesizing perceptually realistic details even when the observation is severely degraded~\cite{li2025diffusion}. 

In particular, diffusion models are especially effective. The formulation of the denoising diffusion process in the image space~\cite{ho2020denoising} or latent space~\cite{rombach2022high}, in combination with the large-scale training of attention-based neural architectures~\cite{vaswani2017attention}, enables diffusion models to be utilized as versatile image priors for various computer vision tasks, including blind image restoration~\cite{he2025diffusion}. Extending these large-scale models for the image restoration task is achieved either by training task-specific architectures from scratch or by including auxiliary feature extractors that condition the pretrained models on degraded observations. While effective, the former sacrifices the strong image prior learned from large-scale data, and the latter incurs substantial training and memory costs due to the large number of additional trained parameters.

Large pretrained diffusion models belong to the broader class of generalizable large-scale self-supervised models, \ie, foundation models~\cite{awais2025foundation}. One prominent example from this class of models, CLIP~\cite{radford2021learning}, has been shown to exhibit a degree of robustness to distribution shifts caused by degradation. Cheng \etal demonstrated that CLIP features remain informative under out-of-distribution noise, enabling effective denoising by training only parameter-efficient decoders~\cite{cheng2024transfer}.

\begin{figure}
    \centering
    \includegraphics[width=0.6\linewidth]{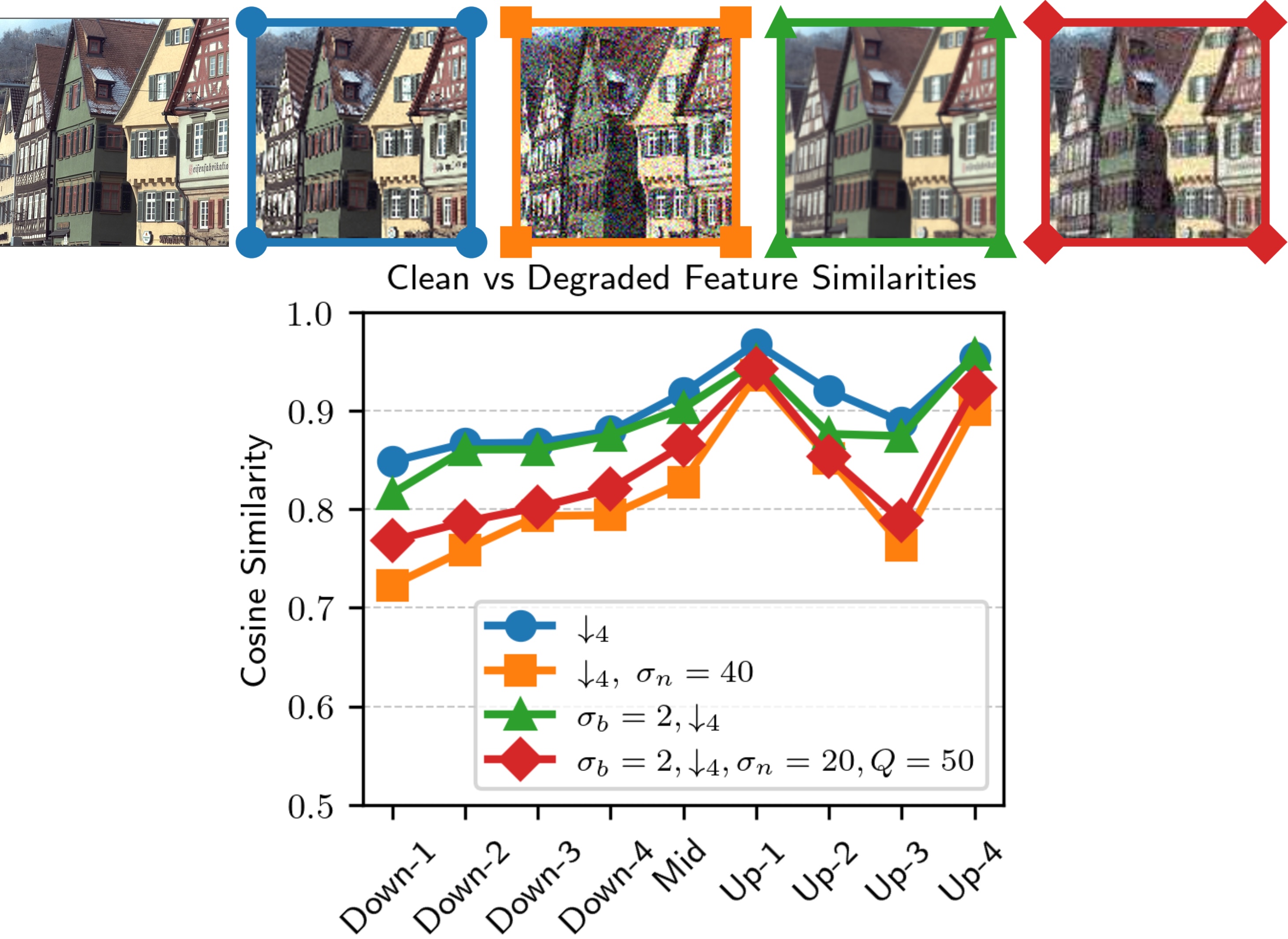}
    \caption{Cosine similarity between latent representations of clean and degraded images across different layers of a U-Net-based latent diffusion model. Similarities are measured at the outputs of the Downsampling (Down), Middle (Mid), and Upsampling (Up) blocks of the U-Net. The degradations are combinations of $4\times$ downsampling ($\downarrow_4$), additive white noise ($\sigma_n$), Gaussian blur ($\sigma_b$), and JPEG compression ($Q$).}
    \label{fig:robust}
\end{figure}

Inspired by this and to provide intuition for our approach, we examine how a pretrained latent diffusion model responds to degraded inputs. As shown in \cref{fig:robust}, we analyze the discrepancy in latent representations by measuring the cosine similarity between latent features of a clean image and its degraded counterpart. The results hint that feature similarity is preserved to varying degrees across layers and degradation types. While this analysis does not constitute proof of robustness, it provides empirical evidence that pretrained diffusion models retain informative representations under degradations. This observation suggests that auxiliary feature extractors, commonly used to guide diffusion-based restoration models, may be unnecessary, and that we can leverage the diffusion model's features for image restoration, enabling parameter-efficient conditioning of pretrained diffusion backbones.

Based on this insight, we introduce \emph{BIR-Adapter}, a parameter-efficient, adapter attention mechanism for diffusion-based blind image restoration. Rather than training an auxiliary network to extract degradation-aware features, BIR-Adapter reuses features extracted by the frozen diffusion backbone itself and integrates them into lightweight restoring attention modules. To further improve inference reliability, BIR-Adapter additionally incorporates a guided sampling strategy that mitigates hallucinations, particularly in low-frequency regions. We evaluate BIR-Adapter on both synthetic and unknown real-world degradation benchmarks. Our results demonstrate that the proposed method achieves competitive and, in several cases, superior performance compared to state-of-the-art approaches, while requiring up to $36\times$ fewer additional trained parameters. Moreover, the adapter-based design enables integration into existing diffusion models. As a proof of concept, we extend a super-resolution–only diffusion model to handle additional unknown degradations, highlighting the flexibility and generality of our approach.

\section{Related work}
\label{sec:related}

\subsection{Deep image restoration}

Image restoration is a long-standing problem in low-level computer vision, with early methods relying on handcrafted priors tailored to specific degradations~\cite{dabov2007image}. With the rise of deep learning, CNN-based models significantly improved restoration performance by learning mappings from degraded images to clean targets, addressing tasks such as denoising~\cite{zhang2017learning}, super-resolution~\cite{dong2015image}, and deblurring~\cite{nah2017deep}. However, these methods typically assumed known or single degradations and showed limited generalization to unknown or compounded cases~\cite{su2022survey}. To mitigate this, later works introduced more realistic degradation modeling~\cite{zhang2021designing} and architectural enhancements, including multi-scale representations~\cite{cui2023irnext} and enlarged receptive fields~\cite{cui2024omni}. Transformer-based models further improved performance by capturing long-range dependencies, as demonstrated by SwinIR~\cite{liang2021swinir}, Restormer~\cite{zamir2022restormer}, and Uformer~\cite{wang2022uformer}. More recent efforts also explored frequency-domain modeling~\cite{kong2023efficient}, subbands~\cite{cui2023selective}, parameter-efficient attention mechanisms~\cite{cui2025modumer}, and frequency-domain all-in-one restoration~\cite{liu2025uhd}. AdaPrompt-IR further extended all-in-one restoration by introducing degradation semantics~\cite{sun2026adaprompt}. Despite these advances, most deep learning–based restoration methods remain regression-driven and struggle to recover perceptually realistic details under severe ambiguity, thereby limiting their ability to recover details in blind settings.

\subsection{Generative image restoration}

Generative image restoration was initially explored using GAN-based methods such as BSRGAN~\cite{zhang2021designing}, and Real-ESRGAN~\cite{wang2021real}, which introduced realistic synthetic degradation pipelines to enable blind restoration under complex and mixed degradations. Generative restoration, in contrast to regression-based approaches, emphasizes perceptual quality rather than strict pixel-level fidelity. While diffusion models do not change this fundamental trade-off, they have been shown to outperform GANs in image synthesis~\cite{dhariwal2021diffusion}. Moreover, large-scale pretrained diffusion models have emerged as particularly strong priors over natural images, making them well-suited for generative image restoration~\cite{li2025diffusion}.

ControlNet introduced a way to steer these powerful priors by fine-tuning a pretrained copy of the model that injects features into the original model~\cite{zhang2023adding}. This prior-based control is also applied to restoration. PASD introduced pixel-wise cross-attention and a ControlNet-based super-resolution module~\cite{yang2024pixel}, while StableSR introduced feature warping to improve image restoration~\cite{wang2024exploiting}. Despite their effectiveness, such models still face challenges when dealing with unknown or compounded degradations. DiffBIR addressed this by incorporating pretrained restoration models, such as SwinIR, into a ControlNet-based pipeline to better generalize across degradation types~\cite{lin2024diffbir}. SUPIR extended this two-stage approach to large-scale degradation estimation~\cite{yu2024scaling}, and Kong \etal extended it with Transformers~\cite{kong2025dual}. AdaptBIR divided the degradation estimation into levels via quality estimation~\cite{liu2024adaptbir}. Luo \etal used visual instructions for higher-level degradation description for diffusion-based all-in-one restoration~\cite{luo2025visual}. Similarly, task-based restoration was utilized by SeeSR~\cite{wu2024seesr} and EDTR~\cite{kim2025exploiting}. Though very effective at restoring perceptually high-quality and realistic images, these approaches require a feature extractor, namely ControlNet.

\subsection{Parameter-efficient diffusion models}

To address the high complexity associated with large ControlNet-based models, Cheng \etal proposed lightweight architectures trained from scratch for super-resolution, introducing DiT-SR~\cite{cheng2025effective}. However, this eliminated the strong prior associated with large diffusion models. One efficient approach is low-rank adaptation~\cite{hu2021lora}, which is not applicable to image restoration because it assumes the model is static, whereas restoration models need to be extended to account for degraded images. On the other hand, adapter-based methods emerged as a low-complexity alternative to full fine-tuning~\cite{ma2025efficient}. Liang \etal introduced the Diffusion Restoration Adapter, which combined adapters and lower rank training as a low-complexity approach to blind image restoration~\cite{liang2025diffusion}. However, similar to ControlNet, adapter-based methods, although more parameter-efficient, still rely on external feature extractors. As discussed in the previous section, the diffusion prior itself removes the need for such auxiliary modules. Leveraging this insight, we introduce BIR-Adapter, illustrated in \Cref{fig:pipeline}.

\section{Background}

In this section, we provide the necessary background that underpins our work. We divided the background into three main parts, in which we discuss the latent diffusion models, sampling, and attention mechanisms.

\subsection{Latent diffusion models}

Diffusion models aim to revert a known $T$-step forward diffusion process defined by the decaying scheduling parameters $\beta_t \quad \text{where } t \in \{0, \dots, T \}$~\cite{ho2020denoising}. This process begins with a known vector $\mathbf{x_0}$ and gradually introduces noise up to timestep $t$. This defines a Markovian forward process $q$ as 
\begin{equation}
    q(\mathbf{x}_t \mid \mathbf{x}_{t-1}) = \mathcal{N}(\mathbf{x}_t;\sqrt{1 - \beta_t}\mathbf{x}_{t-1},~\beta_t \mathbf{I})\text{.}
\end{equation}

We can compute the noisy state $\mathbf{x_t}$ by the reparameterization trick where $\bar{\alpha}_t$ represents the cumulative scheduling parameter and formulated as $\bar{\alpha}_t = \prod_{i=0}^{i=t} \alpha_i$ where $\alpha_t = 1 - \beta_t$

\begin{equation} \label{eq:forward}
    q(\mathbf{x}_t \mid \mathbf{x}_0) = \mathcal{N}(\mathbf{x}_t; \sqrt{\bar{\alpha}_t} \mathbf{x}_0, (1-\bar{\alpha}_t) \mathbf{I}) \text{.}
\end{equation}

A diffusion model aims to reverse this process by modeling the reverse process as a parametric function $p_\theta$

\begin{equation}
    p_\theta(\mathbf{x}_{t-1} \mid \mathbf{x}_t) = \mathcal{N}\big(\mathbf{\mu}_\theta(\mathbf{x}_t, t),~\mathbf{\Sigma}_\theta(\mathbf{x}_t, t)\big)\text{,}
\end{equation} where $\mathbf{\mu}_\theta$ and $\mathbf{\Sigma}_\theta$ are controlled by a denoising network $\epsilon_\theta(\mathbf{x}_t, t)$. This model is trained to minimize

\begin{equation}
    \mathbb{E}\big[(\boldsymbol{\epsilon} - \epsilon_\theta(\sqrt{\bar{\alpha}_t}\mathbf{x}_0 + \sqrt{1-\bar{\alpha}_t}\epsilon, t))^2\big] \text{,}
    \label{eq:loss}
\end{equation} where the noisy data is generated using the forward model in \cref{eq:forward} reformulated as 

\begin{equation} \label{eq:reparameterized}
\mathbf{x}_t
= \sqrt{\bar{\alpha}_t}\,\mathbf{x}_0
+ \sqrt{1-\bar{\alpha}_t}\,\boldsymbol{\epsilon},
\qquad
\boldsymbol{\epsilon} \sim \mathcal{N}(0,~\mathbf{I}).
\end{equation}

Although diffusion models have been shown to outperform GANs, stable synthesis remained challenging~\cite {dhariwal2021diffusion}. Rombach \etal resolved this issue by introducing the latent diffusion models (LDM)~\cite{rombach2022high}. Instead of supplying a noise-free image $I_0$ as $\mathbf{x}_0$, they perform the diffusion process on a latent space, dictated by an autoencoder with an encoder $\mathbf{x}_0 = \mathcal{E}(I_0)$ and a decoder $I_0 = \mathcal{D}(\mathbf{x}_0)$.

Finally, for controllable diffusion models, the denoising model is extended to include a control signal $\mathbf{y}_0$ as $\epsilon_\theta(\mathbf{x}_t, \mathbf{y}_0, t)$. This conditioning on $\mathbf{y}_0$ is achieved by training with $\mathbf{y}_0$ from scratch or by external conditioning such as ControlNet~\cite{zhang2023adding} or adapters~\cite{ma2025efficient}. The former requires large datasets to train billion-scale models from scratch, while the latter enables using these hyper-scale LDMs as priors. ControlNet fine-tunes a pretrained copy of $\epsilon_\theta$ to extract features from $\mathbf{y}_0$ and inject them into the original model. However, as discussed above, this approach also requires fine-tuning of a large set of parameters, thereby increasing the training complexity. Adapter-based approaches extract features from $\mathbf{y}_0$ more efficiently and, instead of using residual connections, they utilize cross-attention at the layers of the original model. This approach reduces complexity; however, it still requires feature extraction. In \Cref{sec:intro}, we empirically show that auxiliary feature extraction is inefficient for diffusion-based blind image restoration, as the diffusion model’s latent representations for clean and degraded inputs exhibit similarity to some degree.

\subsection{Sampling} \label{sec:sampling}

During inference, the goal is to go from noise $\mathbf{x}_T \sim \mathcal{N} (0, \mathbf{I})$ to a clean image $\mathbf{x}_0$. Song \etal introduced Denoising Diffusion Implicit Models, which utilize the estimated noise-free latents $\mathbf{x}_{t\rightarrow0}$ to solve the inverse process

\begin{equation}
    \mathbf{x}_{t-1} = \sqrt{\bar{\alpha}_{t-1}} \mathbf{x}_{t\rightarrow0} + \sqrt{1-\bar{\alpha}_{t-1}}\epsilon_\theta(\mathbf{x}_t, \mathbf{y}_0, t) \text{.}
    \label{eq:reverse}
\end{equation}

The clean latents $\mathbf{x}_{t\rightarrow0}$ are estimated by simply inverting the forward model in \cref{eq:reparameterized} and replacing the initial noise with the denoiser

\begin{equation}
    \mathbf{x}_{t\rightarrow0} = \bigg(\frac{\mathbf{x}_t - \sqrt{1-\bar{\alpha}_t} \epsilon_\theta(\mathbf{x}_t, \mathbf{y}_0, t)}{\sqrt{\bar{\alpha}_t}}\bigg) \text{.}
    \label{eq:estimate}
\end{equation}

\subsection{Attention mechanism}

\begin{figure*}
    \centering
    \begin{subfigure}[b]{0.70\textwidth}
        \centering
        \includegraphics[width=\linewidth]{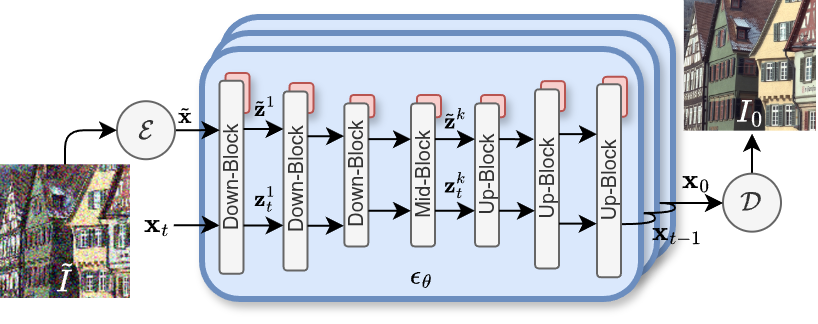}
        \caption{Latent diffusion model with BIR-Adapter}
        \label{fig:denoiser}
    \end{subfigure}
    \\
    \begin{subfigure}[b]{0.35\textwidth}
        \centering
        \includegraphics[width=\linewidth]{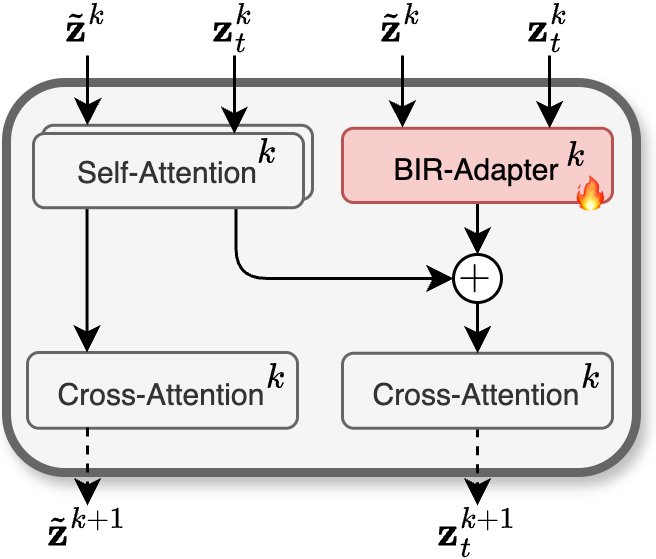}
        \caption{Restoring attention}
        \label{fig:block}
    \end{subfigure}
    \hspace{1pt}
    \begin{subfigure}[b]{0.32\textwidth}
        \centering
        \includegraphics[width=\linewidth]{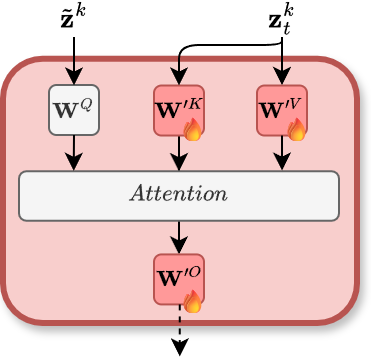}
        \caption{BIR-Adapter}
        \label{fig:bir}
    \end{subfigure}
    \caption{BIR-Adapter in a denoising diffusion model $\epsilon_\theta$. BIR-Adapter (\subref{fig:bir}) introduces a restoring attention mechanism within an attention block (\subref{fig:block}) of the denoising model (\subref{fig:denoiser}). The model incorporates intermediate features $\tilde{\mathbf{z}}^k$ of the degraded latents $\tilde{\mathbf{x}}$, which are extracted by $\epsilon_\theta$ itself. This design leverages the observation that diffusion models retain informative representations under degradation and enables a parameter-efficient adaptation without auxiliary feature extractors. For clarity, the U-Net–based LDM in (\subref{fig:denoiser}) is simplified by visualizing fewer blocks and omitting residual connections, and the number of attention layers is reduced. In (\subref{fig:block}), self-attention is applied in a cascaded manner with parallel processing of degraded and diffused features.}
    \label{fig:pipeline}
\end{figure*}

Attention, introduced in the Transformer architecture~\cite{vaswani2017attention}, allows models to dynamically weight input features based on their relevance to each other. This is achieved through learned similarity scores between query and key vectors. Let $\mathbf{z}^k_t$ be the output of the layer $k$ of $\epsilon_\theta(\mathbf{x}_t, \mathbf{y}_0, t)$. Self-attention is computed as 

\begin{equation}
    Attention(Q,K,V) = softmax(\frac{QK^T}{\sqrt{d_k}})V
\end{equation} where $Q = \mathbf{z}^k_t \mathbf{W}^Q$, $K = \mathbf{z}^k_t \mathbf{W}^K$, and $V = \mathbf{z}^k_t \mathbf{W}^V$, and $\{\mathbf{W}^Q, \mathbf{W}^K, \mathbf{W}^V\}$ constitute the trainable attention weights. Key ($K$) and value ($V$) can be computed via some other modality $\mathbf{y}_0$ to form cross-attention, where $K = \mathbf{y}_0 \mathbf{W}^K$, and $V = \mathbf{y}_0 \mathbf{W}^V$. A typical LDM comprises self-attention and cross-attention layers arranged consecutively, as shown in \cref{fig:pipeline}. Both attention types finally transform the attention output via $\mathbf{W}^O$.

\section{Methodology}

In blind image restoration, an unknown degradation function $\mathcal{F}$ and noise $\eta$ were applied to an image such that $\tilde{I} = \mathcal{F}(I_0) + \eta$, and the goal is to restore $I_0$ from $\tilde{I}$. We introduce the BIR-Adapter, which comprises a restoring attention-based adapter for LDMs and a diffusion guidance mechanism. As motivated in \cref{sec:intro} we utilize LDMs as natural image priors for parameter efficiency. Since we utilize LDMs, we treat this problem as restoration on the latent space, \ie obtaining $\mathbf{x}_0$ from degraded latents $\tilde{\mathbf{x}} = \mathcal{E}(\tilde{I})$.

\subsection{BIR-Adapter}
\label{sec:bir_adapter}
Intuitively, attention determines how query features aggregate information from value features based on key–query relevance. In blind image restoration, this corresponds to refining degraded features based on their relevance to the clean image. This ideal formulation would require access to clean features as keys and values, which are unavailable in the blind setting.

Parallel to the diffused latents $\mathbf{x}_t$, we pass the degraded latents $\tilde{\mathbf{x}}$ through $\epsilon_\theta$ in the same batch and end up with, for each layer $k$, the features $\tilde{\mathbf{z}}^k$. As we discussed in \cref{sec:sampling}, the reverse diffusion process solved by a denoising model $\epsilon_\theta$, denoises, \ie cleans an image. Inspired by this, we discussed in \cref{sec:intro} that one can form the hypothesis that when $\epsilon_\theta$ is pretrained at scale, degraded features should exhibit similarity to clean ones. \cref{fig:robust} shows, for an example image under various degradations, a proof of concept for this idea. Please note that we introduced \Cref{fig:robust} as a visual motivation and do not claim the diffusion models' general robustness. Factors such as the diffusion timestep and the text prompt affect the latent representations. However, the degraded features should contain enough information about the original image, as there are similarities between the features $\mathbf{z}^k_t$ and $\tilde{\mathbf{z}}^k$. As such, we treat the currently diffused features $\mathbf{z}^k_t$ of a layer $k$ at diffusion timestep $t$ as the \textit{clean} features that were missing in the formulation above. Using these definitions, we introduce the restoring attention mechanism as follows

\begin{equation}
\begin{aligned}
    & Attention(Q,K,V)\mathbf{W}^O + Attention(\tilde{Q}, K', V')\mathbf{W}'^O \text{,}
\end{aligned}
\end{equation} where $\tilde{Q} = \tilde{\mathbf{z}}^k \mathbf{W}^Q$, $K' = \mathbf{z}^k_t \mathbf{W}'^K$, and $V' = \mathbf{z}^k_t \mathbf{W}'^V$. This formulation allows us to fine-tune the parameter set $\{\mathbf{W}'^K, \mathbf{W}'^V, \mathbf{W}'^O\}$ of each self-attention layer, initialized via their self-attention counterparts, resulting in parameter efficiency. \Cref{fig:pipeline} visualizes this architecture inside a simplified LDM. We do not fine-tune the query weights since we want to preserve the features of the degraded image.

\subsection{Guided sampling} \label{sec:guidance}

As an image's resolution increases and exceeds the supported resolution of an LDM, a common practice is to tile the latent space with overlaps and run the diffusion on the tiles. Finally, the tiles are merged with certain weighting functions~\cite{wang2024exploiting}. In our approach, we observed that if a tile contains mainly low-frequency information, BIR-Adapter leads to some severe hallucinations. A common practice to avoid this issue is to use an initial restoration step, either via an off-the-shelf method~\cite{lin2024diffbir} or a custom-trained one~\cite{yu2024scaling}, and combine them with frequency estimation~\cite{kim2025exploiting}. We assume an initial restoration $I_0^{init}$ exists.

We use this information to guide the diffusion-based sampling discussed in \cref{sec:sampling}. However, we do not apply this guidance to high-frequency regions. To achieve this, we utilize the downsampled and normalized gradients $\mathcal{G}_\downarrow$ estimated via the Sobel filter~\cite{lin2024diffbir}. We would like to note that the filter choice does not affect the results, as we are only searching for a general estimate. We define the downsampled and normalized gradient operator $\mathcal{G}_\downarrow(I)$ applied to an input image \(I\) as

\begin{equation}
    \mathcal{G}_\downarrow(I) = \frac{\text{Sobel}(I) \downarrow s}{\|\text{Sobel}(I) \downarrow s\|} \text{,}
\end{equation} where \(\downarrow s\) indicates downsampling by a factor \(s\), and division with the norm \(\|\cdot\|\) normalizes the result to the range \([0,1]\). Here, the downsampling factor $s$ is dictated by the encoder $\mathcal{E}$ of the latent diffusion model. Using this operator, we compute the corresponding guidance weights as adapted from DiffBIR~\cite{lin2024diffbir} as

\begin{equation}
    \mathcal{W} = 1 - \mathcal{G}_\downarrow(I_0^{init}) \text{.}
\end{equation}
    
We then use the estimated noise-free latent $\mathbf{x}_{t\rightarrow0}$ found in \cref{eq:reverse} and, for $\tfrac{t}{T}>\xi$, update it to include the clean low-frequency information to reduce the likelihood of hallucination

\begin{figure}[t]
    \centering
    \begin{subfigure}[b]{1\textwidth}
        \begin{subfigure}[b]{0.25\linewidth}
            \centering
            \includegraphics[width=\linewidth]{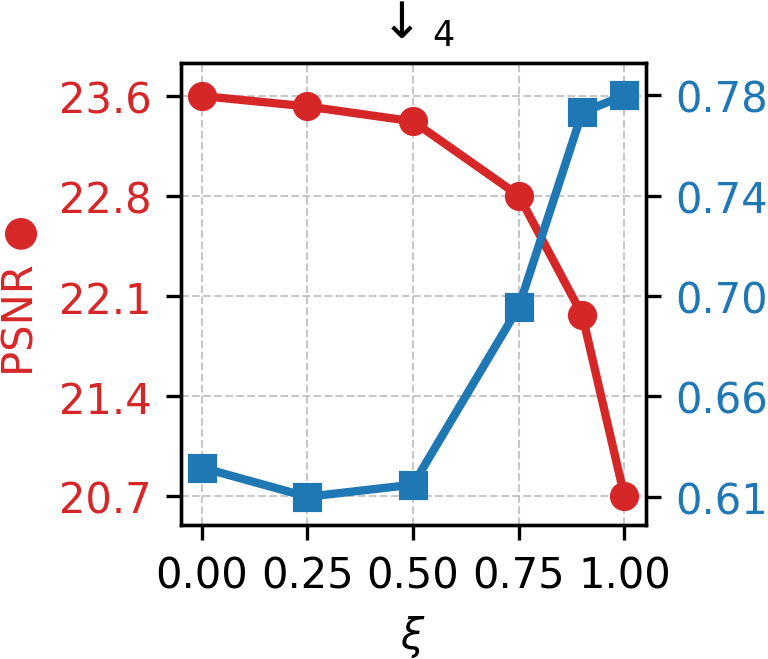}
        \end{subfigure}
        \hfill
        \begin{subfigure}[b]{0.23\linewidth}
            \centering
            \includegraphics[width=\linewidth]{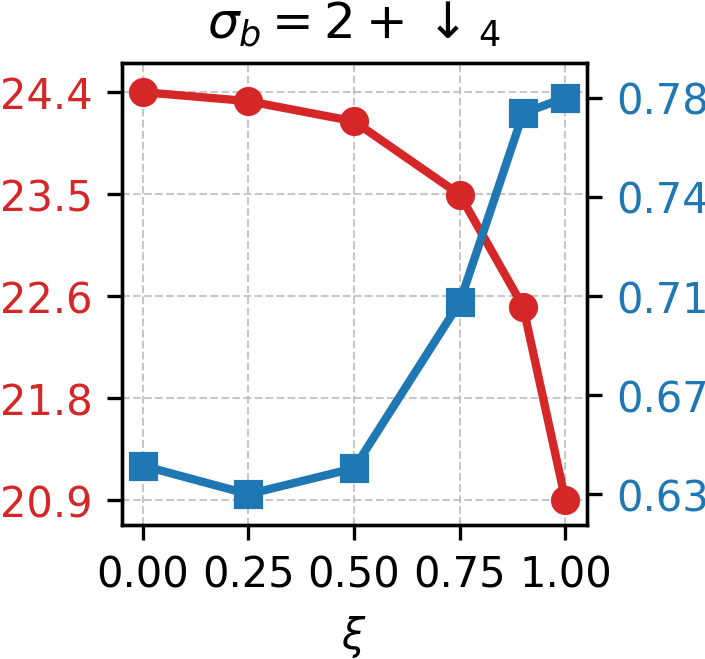}
        \end{subfigure}
        \hfill
        \begin{subfigure}[b]{0.23\linewidth}
            \centering
            \includegraphics[width=\linewidth]{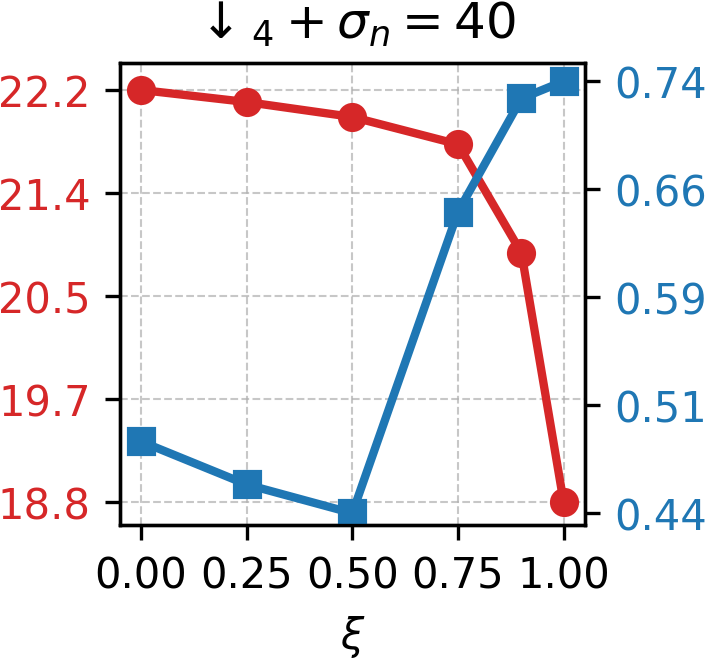}
        \end{subfigure}
        \hfill
        \begin{subfigure}[b]{0.256\linewidth}
            \centering
            \includegraphics[width=\linewidth]{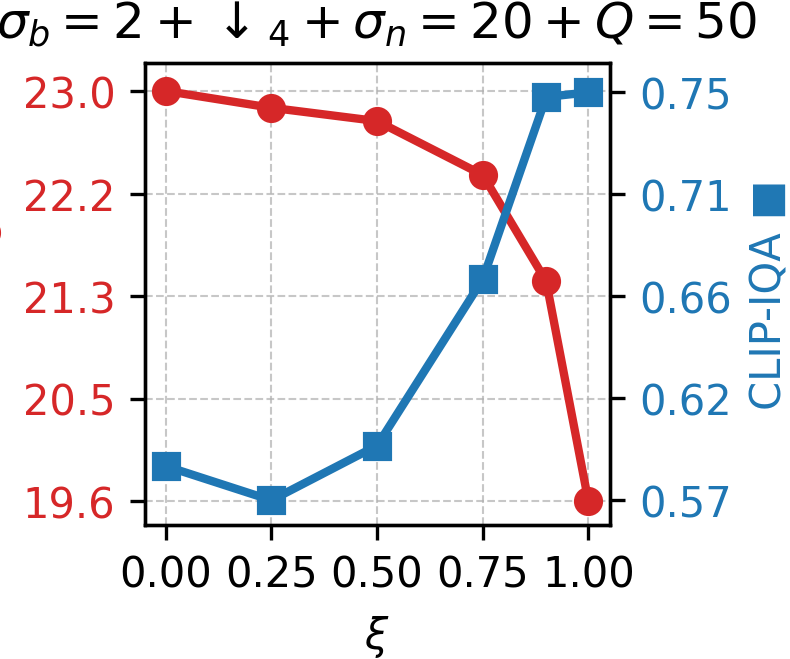}
        \end{subfigure}
    \caption{Quantitative effect of $\xi$ in terms of PSNR and CLIP-IQA under various combinations of 4$\times$ downsampling ($\downarrow_4$), white noise ($\sigma_n$), Gaussian blur ($\sigma_b$), and JPEG ($Q$). Reported values are averaged over the DIV2K validation dataset.}
    \label{fig:xi_numerical}
    \end{subfigure}
    \\
    \begin{subfigure}[b]{1\linewidth}
        \centering
        \includegraphics[width=\linewidth]{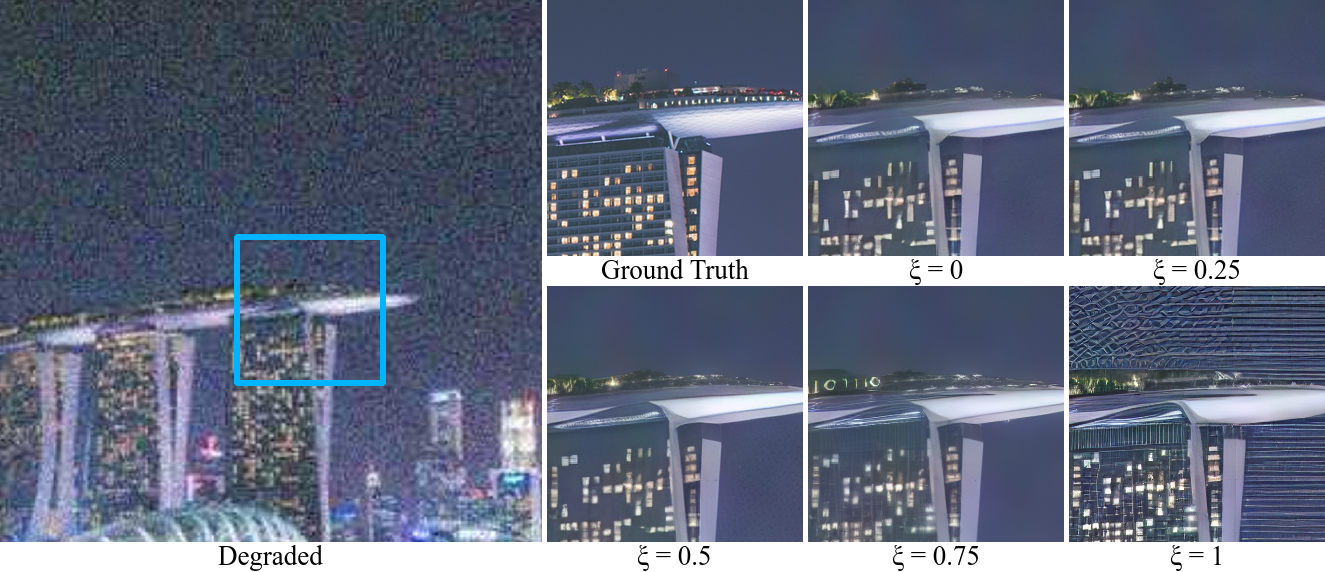}
        \caption{Visual effect of the guidance parameter $\xi$ on the restoration of a single example.}
        \label{fig:w_wo_guidance}
    \end{subfigure}
    \caption{Quantitative and visual analysis of the effect of $\xi$. An increase in CLIP-IQA indicates higher-quality images, while a sudden drop in PSNR suggests inconsistencies.}
    \label{fig:guidance}
\end{figure}

\begin{equation}
    \mathbf{x}_{t\rightarrow0} \leftarrow 
    \begin{cases}
        \mathbf{x}_{t\rightarrow0} + \mathcal{W} \odot \big(\mathcal{E}(I_0^{init}) - \mathbf{x}_{t\rightarrow0}\big), & \text{if } \frac{t}{T} > \xi \\
        \mathbf{x}_{t\rightarrow0}, & \text{o/w.}
    \end{cases}
\end{equation}

\Cref{fig:guidance} provides quantitative and visual analysis of the adapted guidance mechanism, with quantitative results computed over the evaluation dataset. We examine the effect of $\xi$ on fidelity, measured by PSNR, and perceptual quality, measured by CLIP-IQA~\cite{wang2022exploring}. Stronger guidance (lower $\xi$) increases PSNR but reduces perceptual quality, while weaker guidance (higher $\xi$) improves CLIP-IQA at the cost of PSNR. Notably, disabling guidance ($\xi = 1$) results in a sharp drop in PSNR, indicating increased inconsistency. Overall, $0.75 \leq \xi \leq 0.9$ yields the best trade-off. We emphasize that, for generative restoration methods, a higher PSNR does not necessarily imply better perceptual restoration quality; similarly, a higher CLIP-IQA does not imply consistency. Therefore, we explicitly report the trade-off between PSNR and CLIP-IQA.

\section{Experiments}

We validated our approach through a series of experiments involving simulated and real-world degradations. In this section, we provide the details of the implementation and evaluation setup.

\subsection{Training}

We utilized the Stable Diffusion 1.5 (SD-1.5) backbone. It has $860 \times 10^6$ frozen parameters, and BIR-Adapter introduces only an additional $37 \times 10^6$ extra trainable parameters. We initialized $\mathbf{W}'^K$ and $\mathbf{W}'^V$ from their counterparts and $\mathbf{W}'^O$ as zeros; thus, BIR-Adapter has no initial effect. We discussed how we utilized these weights in \cref{sec:bir_adapter}. \cref{fig:pipeline} illustrates the architecture.

For training, we utilized a mixture of datasets, consisting of the training sets of DIV2K~\cite{agustsson2017ntire}, DIV8K~\cite{gu2019div8k}, Flickr2K~\cite{Lim_2017_CVPR_Workshops}, WED, OutdoorSceneTrain~\cite{wang2018recovering}, and the first 5,000 images from the FFHQ dataset~\cite{karras2019style}, creating a training set of size 25,000 images. During training, we selected a random square patch of size $512 \times 512$. We degraded the patch using the stochastic degradation model utilized by Real-ESRGAN~\cite{wang2021real}. 

Once we have the clean and degraded image pair $I_0$ and $\tilde{I}$, we encode them to the latent space by the encoder of the latent diffusion model $\mathcal{E}$, resulting in $\mathbf{x}_0$ and $\tilde{\mathbf{x}}$ respectively. For each training step, we sampled an $\mathbf{x}_t$ by using the formulation in \cref{eq:forward}. We passed $\tilde{\mathbf{x}}$ and $\mathbf{x}_t$ together to the denoising model $\epsilon_\theta$ and executed the BIR-Adapter-extended model as shown in \cref{fig:pipeline}. We finally used the loss formulation in \cref{eq:loss} to train the parameters of the BIR-Adapter. Notice how $\tilde{\mathbf{x}}$ is always the same for each diffusion step, and it is not utilized in the loss. 

We utilized the AdamW optimizer with learning rate $10^{-5}$ and weight decay of $10^{-2}$~\cite{loshchilov2017decoupled}. We used $3 \times$ NVIDIA A40 GPUs with 48GB VRAM each. We set the batch size and gradient accumulation to $4$. This resulted in an effective batch size of $3 \times 4 \times 4 = 48$. We trained the model for $150 \times 10^3$ steps, which took a week to complete.

\subsection{Evaluation}

\begin{figure*}
    \centering
    \begin{subfigure}[b]{1\linewidth}
        \centering
        \includegraphics[width=\linewidth]{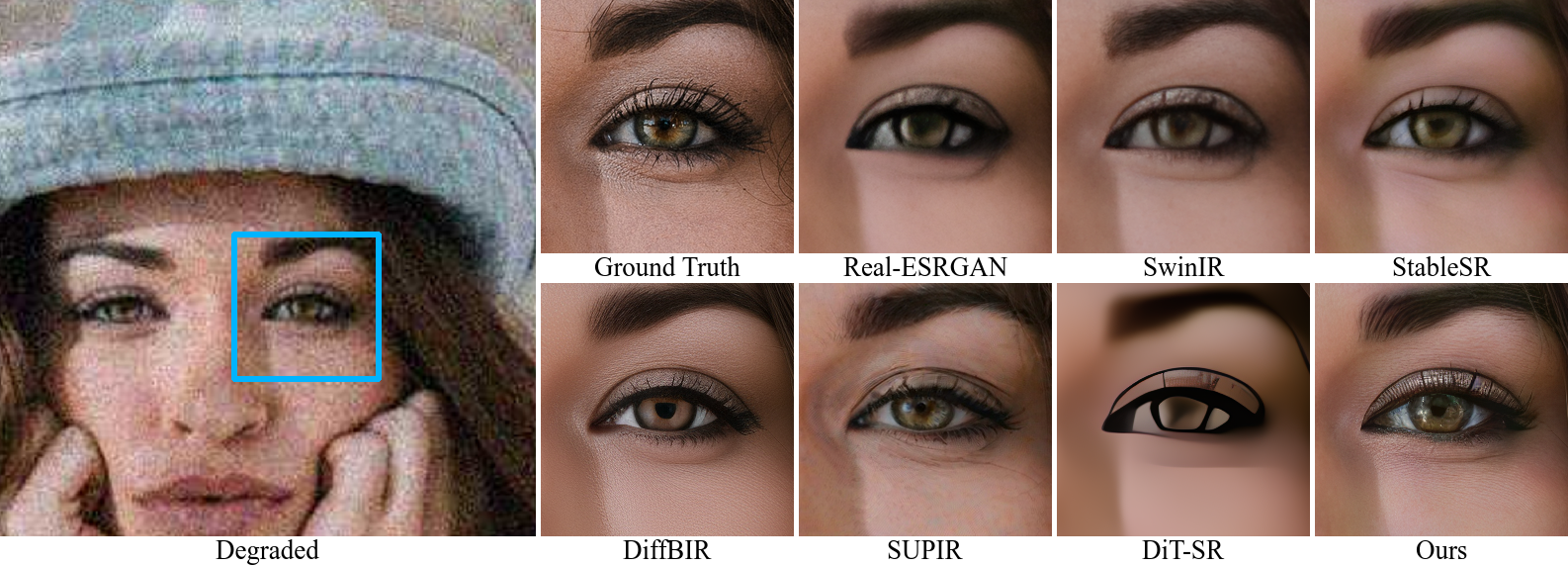}
        \caption{DIV2K: Synthetic degradation $\sigma_{b}=2 + \downarrow_4 + \sigma_{n}=20 + Q=50$.}
        \label{fig:div2k}
    \end{subfigure}

    \begin{subfigure}[b]{1\linewidth}
        \centering
        \includegraphics[width=\linewidth]{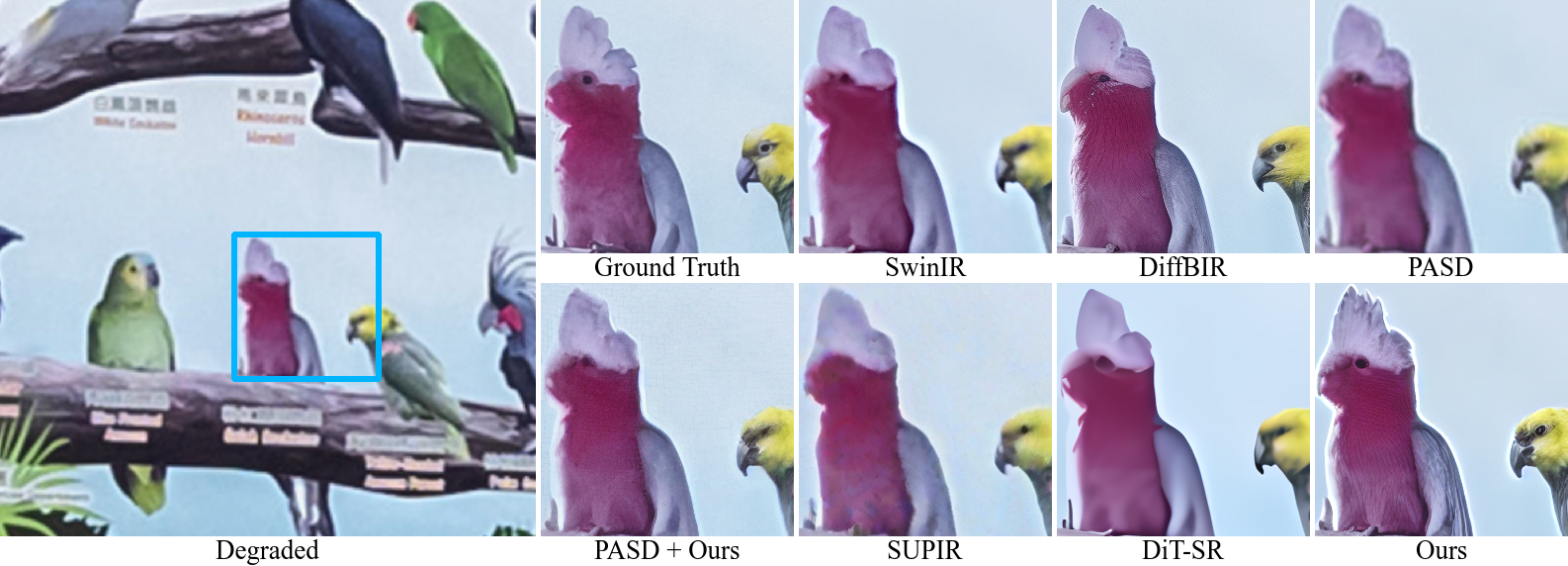}
        \caption{RealSR: unknown real degradation. Restoration of $\downarrow_4 $ and unkown degradations.}
        \label{fig:realsr}
    \end{subfigure}
    \caption{Example degraded and restored images using the baselines and our method. We used synthetic degradation on the DIV2K dataset (\ref{fig:div2k}) while RealSR contains $4 \times$ downsampled images with further unknown degradations (\ref{fig:realsr}).}
    \label{fig:qual}
\end{figure*}

We utilized two main datasets to present both quantitative and qualitative results. First, we used the validation set of the DIV2K dataset. We measured the performance on various synthetic degradation settings: (i) downsampling ($\downarrow$), (ii) Gaussian blur and downsampling ($\sigma_b + \downarrow$), (iii)  downsampling and white noise ($\downarrow + \sigma_n$), and (iv) Gaussian blur, downsampling, white noise, and JPEG compression ($\sigma_b + \downarrow + \sigma_n + Q$). These settings represent common and challenging mixed degradations in blind restoration. We further evaluated our method using RealSR~\cite{cai2019toward}. While degradations such as motion blur or spatially varying effects are not explicitly modeled in the synthetic setting, the RealSR dataset includes unknown real-world degradations. \Cref{fig:qual} displays examples from both datasets. We always included super-resolution as a part of the restoration of these two datasets. We tiled the latent space to a size of $64 \times 64$ with a stride of $32 \times 32$ to support high-resolution images. We combined these tiles after each diffusion step using Gaussian weights and then re-tiled them. This iterative approach avoids blocking~\cite{wang2024exploiting}.

We ran the synthesis for 20 diffusion steps, which provides a good balance between restoration quality and computational cost and is commonly adopted in recent diffusion-based image restoration works. We used class-free guidance with a guidance weight of $9.0$~\cite{ho2022classifier}. We set the guidance parameter $\xi = 0.9$. We qualitatively and quantitatively discussed this choice in \cref{sec:guidance}. We used Real-ESRGAN for the initial restoration $I_0^{init}$, which serves as a lightweight and widely used baseline capable of handling complex and mixed degradations, while contributing only marginally to the overall computational cost.

\subsubsection{Baseline methods}

We compared our method with non-generative SwinIR~\cite{liang2021swinir}, GAN-based Real-ESRGAN~\cite{wang2021real}, and diffusion prior-based methods that utilize ControlNet, such as StableSR~\cite{wang2024exploiting} and PASD~\cite{yang2024pixel}, which are specifically designed for super-resolution. The central focus of our method is to address unknown and complex degradations. Hence, we included DiffBIR~\cite{lin2024diffbir} and SUPIR~\cite{yu2024scaling} in our comparisons as universal restoration methods. Finally, we compared our approach with DiT-SR, which does not utilize a pretrained diffusion prior~\cite{cheng2025effective}. We set up the same synthetic degradation experimentation as SUPIR. We used the source codes provided by the authors to run the baseline models with static prompts. We compare the number of additional trained parameters and the number of images used for training in \cref{tab:parameters} and provide qualitative comparisons in \cref{fig:qual}.

\begin{table}[t]
\centering
\caption{Comparison of the number of additional \emph{trainable} parameters, number of training images, and restoration time. The reported parameter counts exclude the frozen backbone parameters, which are nevertheless involved during inference.}
{\tiny
\begin{tabular}{cccccccc}
\hline
 & StableSR & DiffBIR & PASD & SUPIR & DiT-SR & DR-Adapter & Ours \\ \hline
Trainable Parameters ($\times 10^6$)  & 490 & 363 & 609 & 1,332 & 61 & 80 & 37 \\
Training Images ($\times 10^3$)      & 15 & 15,000 & 45 & 20,000 & 98 & 300 & 25 \\
Restoration time (s)      & 3.03 & 3.86 & 3.60 & 7.50 & 4.42 & - & 7.42 \\
\hline
\end{tabular}
}
\label{tab:parameters}
\end{table}

\subsubsection{Parameter efficiency} \label{sec:parameter}

We introduce a parameter-efficient approach that leverages degraded features extracted by the diffusion model itself, rather than relying on auxiliary feature extractors such as ControlNet. \Cref{tab:parameters} compares the number of additional trained parameters of the baseline methods, along with restoration times, measured for $512 \times 512$ inputs using 20 diffusion steps on an A40 GPU. Our approach requires significantly fewer parameters. We note that parameter efficiency does not directly translate to lower inference cost: one limitation of our approach is the increased runtime, as the degraded image is processed in parallel with the diffused sample. We discuss this in more detail below.

\subsubsection{Adapter baseline} \label{sec:adapter_baseline}

We also compare our approach with a recent adapter-based blind image restoration method, namely DR-Adapter, which we discussed in \cref{sec:related}~\cite{liang2025diffusion}. However, the approach is closed-source. The authors reported performance on the DIV2K dataset under synthetic degradations, but the degradation parameters are unknown. However, they report real-life performance on the RealPhoto60 dataset~\cite{yu2024scaling}. We compare the performances on the RealPhoto60 dataset in \cref{tab:real60}, including DR-Adapter performance as reported by the authors. We do not use guidance in this setting, as tiling is not necessary.

\subsubsection{Metrics}

We utilized a wide range of metrics to measure the performance. These metrics include pixel-level reference-based metrics such as PSNR and MS-SSIM, a perceptual reference-based metric LPIPS~\cite{zhang2018perceptual}, and reference-free metrics CLIP-IQA~\cite{wang2022exploring}, ManIQA~\cite{yang2022maniqa}, and MUSIQ~\cite{ke2021musiq} that are shown to correlate with human preferences. The reference-free metrics are shown to correlate more with human perception, as pixel-level reference-based metrics do not reflect the perceptual quality of an image. However, reference-free metrics do not necessarily imply consistency, as we discussed in the context of hallucination-quality trade-off in \cref{sec:guidance}. We report qualitative results in \cref{fig:qual} and additional ones in the appendix. We report performance of the methods in terms of the aforementioned metrics in \cref{tab:div2k}, \cref{tab:realsr}, and \cref{tab:real60}.

\begin{sidewaystable}[htbp]
\centering
{\scriptsize
\begin{tabular}{ccccccccccc}
\toprule
\makecell{Experiment} & \makecell{Metric} & \makecell{Real-ESRGAN \\ (ICCV '21)} & \makecell{SwinIR \\ (ICCV '21)} & \makecell{StableSR \\ (IJCV '24)} & \makecell{DiffBIR \\ (CVPR '24)} & \makecell{PASD \\ (ECCV '24)} & \makecell{SUPIR \\ (CVPR '24)} & \makecell{DiT-SR \\ (AAAI '25)} & \makecell{Ours} & \makecell{Ours \\ w/o guidance} \\ \midrule
\multirow{6}{*}{\shortstack{$\downarrow_4$}} & CLIP-IQA & 0.566 & 0.583 & 0.608 & \cellcolor{gray!20}\textbf{\textcolor{red}{0.788}} & 0.534 & 0.532 & 0.552 & \cellcolor{gray!20}0.772 & \cellcolor{gray!20}\textit{\textcolor{blue}{0.779}} \\
 & ManIQA & 0.367 & 0.378 & 0.373 & \cellcolor{gray!20}\textbf{\textcolor{red}{0.581}} & 0.358 & 0.359 & 0.397 & \cellcolor{gray!20}0.529 & \cellcolor{gray!20}\textit{\textcolor{blue}{0.548}} \\
 & MUSIQ & 63.597 & 64.487 & 64.798 & \cellcolor{gray!20}\textbf{\textcolor{red}{69.83}} & 60.882 & 63.824 & 63.49 & \cellcolor{gray!20}69.212 & \cellcolor{gray!20}\textit{\textcolor{blue}{69.232}} \\
 & LPIPS $\downarrow$ & \cellcolor{gray!20}\textit{\textcolor{blue}{0.243}} & \cellcolor{gray!20}\textbf{\textcolor{red}{0.233}} & \cellcolor{gray!20}0.245 & 0.322 & 0.267 & 0.264 & 0.264 & 0.29 & 0.37 \\
 & PSNR & 23.873 & 24.014 & \cellcolor{gray!20}24.018 & 22.921 & \cellcolor{gray!20}\textit{\textcolor{blue}{24.314}} & 23.589 & \cellcolor{gray!20}\textbf{\textcolor{red}{24.921}} & 22.188 & 20.827 \\
 & MS-SSIM & \cellcolor{gray!20}0.894 & \cellcolor{gray!20}\textit{\textcolor{blue}{0.899}} & 0.894 & 0.855 & 0.881 & 0.863 & \cellcolor{gray!20}\textbf{\textcolor{red}{0.91}} & 0.839 & 0.783 \\
\midrule
\multirow{6}{*}{\shortstack{$\sigma_{b}=2$ \\ $\downarrow_4$}} & CLIP-IQA & 0.57 & 0.581 & 0.582 & \cellcolor{gray!20}\textbf{\textcolor{red}{0.788}} & 0.476 & 0.568 & 0.563 & \cellcolor{gray!20}0.776 & \cellcolor{gray!20}\textit{\textcolor{blue}{0.782}} \\
 & ManIQA & 0.39 & 0.39 & 0.362 & \cellcolor{gray!20}\textbf{\textcolor{red}{0.579}} & 0.326 & 0.38 & 0.403 & \cellcolor{gray!20}0.529 & \cellcolor{gray!20}\textit{\textcolor{blue}{0.554}} \\
 & MUSIQ & 64.811 & 65.269 & 63.169 & \cellcolor{gray!20}\textbf{\textcolor{red}{70.296}} & 56.439 & 66.602 & 63.302 & \cellcolor{gray!20}\textit{\textcolor{blue}{69.617}} & \cellcolor{gray!20}69.51 \\
 & LPIPS $\downarrow$ & \cellcolor{gray!20}\textit{\textcolor{blue}{0.231}} & \cellcolor{gray!20}\textbf{\textcolor{red}{0.215}} & 0.252 & 0.303 & 0.281 & \cellcolor{gray!20}0.247 & 0.267 & 0.276 & 0.363 \\
 & PSNR & 25.149 & \cellcolor{gray!20}25.665 & 25.213 & 23.667 & \cellcolor{gray!20}\textit{\textcolor{blue}{25.811}} & 24.438 & \cellcolor{gray!20}\textbf{\textcolor{red}{26.18}} & 22.778 & 21.072 \\
 & MS-SSIM & \cellcolor{gray!20}0.923 & \cellcolor{gray!20}\textit{\textcolor{blue}{0.933}} & 0.915 & 0.879 & 0.914 & 0.894 & \cellcolor{gray!20}\textbf{\textcolor{red}{0.946}} & 0.864 & 0.8 \\
\midrule
\multirow{6}{*}{\shortstack{$\downarrow_4$ \\ $\sigma_{n}=40$}} & CLIP-IQA & 0.466 & 0.568 & 0.451 & \cellcolor{gray!20}0.655 & 0.585 & 0.532 & 0.51 & \cellcolor{gray!20}\textit{\textcolor{blue}{0.728}} & \cellcolor{gray!20}\textbf{\textcolor{red}{0.74}} \\
 & ManIQA & 0.286 & 0.307 & 0.27 & 0.363 & 0.399 & 0.363 & \cellcolor{gray!20}0.431 & \cellcolor{gray!20}\textit{\textcolor{blue}{0.473}} & \cellcolor{gray!20}\textbf{\textcolor{red}{0.512}} \\
 & MUSIQ & 50.21 & 57.649 & 50.377 & 49.157 & 62.501 & \cellcolor{gray!20}65.0 & 61.042 & \cellcolor{gray!20}\textit{\textcolor{blue}{68.043}} & \cellcolor{gray!20}\textbf{\textcolor{red}{69.686}} \\
 & LPIPS $\downarrow$ & 0.444 & \cellcolor{gray!20}0.404 & 0.436 & 0.648 & 0.449 & \cellcolor{gray!20}\textbf{\textcolor{red}{0.359}} & 0.457 & \cellcolor{gray!20}\textit{\textcolor{blue}{0.399}} & 0.488 \\
 & PSNR & \cellcolor{gray!20}\textit{\textcolor{blue}{22.016}} & 21.725 & \cellcolor{gray!20}\textbf{\textcolor{red}{22.788}} & 18.39 & 20.733 & 21.87 & \cellcolor{gray!20}22.012 & 21.224 & 19.006 \\
 & MS-SSIM & 0.778 & \cellcolor{gray!20}\textbf{\textcolor{red}{0.804}} & \cellcolor{gray!20}0.794 & 0.699 & 0.721 & 0.765 & \cellcolor{gray!20}\textit{\textcolor{blue}{0.798}} & 0.757 & 0.635 \\
\midrule
\multirow{6}{*}{\shortstack{$\sigma_{b}=2$ \\ $\downarrow_4$\\ $\sigma_{n}=20$ \\ $Q=50$}} & CLIP-IQA & 0.556 & 0.572 & 0.481 & \cellcolor{gray!20}\textbf{\textcolor{red}{0.805}} & 0.479 & 0.563 & 0.513 & \cellcolor{gray!20}0.752 & \cellcolor{gray!20}\textit{\textcolor{blue}{0.754}} \\
 & ManIQA & 0.369 & 0.371 & 0.298 & \cellcolor{gray!20}\textbf{\textcolor{red}{0.592}} & 0.328 & 0.374 & 0.453 & \cellcolor{gray!20}0.5 & \cellcolor{gray!20}\textit{\textcolor{blue}{0.524}} \\
 & MUSIQ & 60.471 & 62.186 & 54.195 & \cellcolor{gray!20}\textbf{\textcolor{red}{70.11}} & 56.65 & 65.921 & 62.542 & \cellcolor{gray!20}69.003 & \cellcolor{gray!20}\textit{\textcolor{blue}{69.755}} \\
 & LPIPS $\downarrow$ & \cellcolor{gray!20}\textit{\textcolor{blue}{0.348}} & \cellcolor{gray!20}\textbf{\textcolor{red}{0.328}} & 0.352 & 0.419 & 0.391 & \cellcolor{gray!20}0.35 & 0.462 & 0.374 & 0.469 \\
 & PSNR & \cellcolor{gray!20}23.267 & \cellcolor{gray!20}\textit{\textcolor{blue}{23.328}} & \cellcolor{gray!20}\textbf{\textcolor{red}{24.152}} & 22.272 & 23.073 & 22.242 & 22.467 & 21.774 & 19.832 \\
 & MS-SSIM & \cellcolor{gray!20}0.833 & \cellcolor{gray!20}\textit{\textcolor{blue}{0.836}} & \cellcolor{gray!20}\textbf{\textcolor{red}{0.839}} & 0.797 & 0.812 & 0.78 & 0.811 & 0.78 & 0.68 \\
\bottomrule
\end{tabular}
}
\caption{Evalution results on dataset DIV2K under various combinations of synthetic degragations of 4$\times$ downsampling ($\downarrow_4$), white noise ($\sigma_n$), Gaussian blur ($\sigma_b$), and JPEG compression ($Q$). We highlight \textbf{\textcolor{red}{best}}, \textit{\textcolor{blue}{second-best}}, and \colorbox{gray!20}{top-3} performances. Lower LPIPS values signify better results, denoted by $\downarrow$. We present the performance of our method both with ($\xi=0.9$) and without guidance ($\xi=1$).}
\label{tab:div2k}
\end{sidewaystable}
\begin{sidewaystable}
\centering
{\scriptsize
\begin{tabular}{ccccccccccc}
\toprule
\makecell{Metric} & \makecell{Real-ESRGAN \\ (ICCV '21)} & \makecell{SwinIR \\ (ICCV '21)} & \makecell{StableSR \\ (IJCV '24)} & \makecell{DiffBIR \\ (CVPR '24)} & \makecell{PASD \\ (ECCV '24)} & \makecell{SUPIR \\ (CVPR '24)} & \makecell{DiT-SR \\ (AAAI '25)} & \makecell{Ours} & \makecell{Ours \\ w/o guidance} \\ \midrule
CLIP-IQA & 0.49 & 0.469 & 0.505 & \cellcolor{gray!20}\textit{\textcolor{blue}{0.751}} & 0.38 & 0.517 & 0.529 & \cellcolor{gray!20}0.741 & \cellcolor{gray!20}\textbf{\textcolor{red}{0.767}} \\
ManIQA & 0.367 & 0.343 & 0.351 & \cellcolor{gray!20}\textit{\textcolor{blue}{0.58}} & 0.295 & 0.386 & 0.508 & \cellcolor{gray!20}0.576 & \cellcolor{gray!20}\textbf{\textcolor{red}{0.609}} \\
MUSIQ & 59.668 & 59.595 & 58.173 & \cellcolor{gray!20}\textit{\textcolor{blue}{67.188}} & 47.608 & 62.35 & 64.484 & \cellcolor{gray!20}\textbf{\textcolor{red}{68.265}} & \cellcolor{gray!20}67.173 \\
LPIPS $\downarrow$ & \cellcolor{gray!20}0.273 & \cellcolor{gray!20}\textbf{\textcolor{red}{0.251}} & \cellcolor{gray!20}\textit{\textcolor{blue}{0.272}} & 0.414 & 0.286 & 0.343 & 0.326 & 0.363 & 0.469 \\
PSNR & \cellcolor{gray!20}24.446 & \cellcolor{gray!20}\textit{\textcolor{blue}{25.02}} & 24.345 & 23.494 & \cellcolor{gray!20}\textbf{\textcolor{red}{26.057}} & 24.126 & 22.658 & 22.143 & 20.347 \\
MS-SSIM & \cellcolor{gray!20}0.886 & \cellcolor{gray!20}\textbf{\textcolor{red}{0.895}} & 0.879 & 0.855 & \cellcolor{gray!20}\textit{\textcolor{blue}{0.89}} & 0.851 & 0.87 & 0.828 & 0.76 \\
\bottomrule
\end{tabular}
\caption{Evaluation results on dataset RealSR, containing $4 \times$ downsampling and unknown degradations. We present the performance of our method both with ($\xi=0.9$) and without guidance ($\xi=1$)} \label{tab:realsr}

\bigskip\bigskip

\begin{tabular}{ccccccccc}
\toprule
\makecell{Metric} & \makecell{SwinIR \\ (ICCV '21)} & \makecell{StableSR \\ (IJCV '24)} & \makecell{DiffBIR \\ (CVPR '24)} & \makecell{PASD \\ (ECCV '24)} & \makecell{SUPIR \\ (CVPR '24)} & \makecell{DR-Adapter \\ (Reported)} & \makecell{Ours \\ w/o guidance} \\ \midrule
CLIP-IQA & 0.646 & 0.523 & \cellcolor{gray!20}\textbf{\textcolor{red}{0.817}} & 0.637 & \cellcolor{gray!20}0.798 & 0.706 & \cellcolor{gray!20}\textit{\textcolor{blue}{0.811}} \\
ManIQA & 0.551 & 0.287 & \cellcolor{gray!20}\textbf{\textcolor{red}{0.646}} & 0.402 & \cellcolor{gray!20}\textit{\textcolor{blue}{0.642}} & 0.529 & \cellcolor{gray!20}0.618 \\
MUSIQ & 57.459 & 31.173 & \cellcolor{gray!20}\textit{\textcolor{blue}{71.946}} & 63.903 & 71.109 & \cellcolor{gray!20}71.62 & \cellcolor{gray!20}\textbf{\textcolor{red}{73.141}} \\
\bottomrule
\end{tabular}
}
\caption{Evaluation results on dataset RealPhoto60, containing unknown degradations. The performance of Diffusion Restoration Adapter (DR-Adapter) is directly reported as the code is not available \cite{liang2025diffusion}. We omit guidance as we do not supersample in this experiment. We highlight \textbf{\textcolor{red}{best}}, \textit{\textcolor{blue}{second-best}}, and \colorbox{gray!20}{top-3} performances. Lower LPIPS values signify better results, denoted by $\downarrow$.} \label{tab:real60}
\end{sidewaystable}

\subsection{Plug \& play}
We selected PASD to evaluate the plug-and-play nature of the BIR Adapter. The reasoning is twofold: (i) PASD shares the same backbone, and (ii) it is super-resolution-only, and we want to show that the performance on unknown degradations can be improved. This experiment serves as proof of concept that our method could enable new applications. We disabled the guidance mechanism as PASD already serves as a guiding restoration. We report the performance of PASD extended with BIR-Adapter on RealSR in \cref{tab:pasd_bir}.

\begin{table}[t]
\centering
    {\tiny
    \begin{tabular}{ccccccc}
    \toprule
    & CLIP-IQA & ManIQA & MUSIQ & LPIPS $\downarrow$ & PSNR & MS-SSIM \\ \midrule
    PASD & 0.38 & 0.295 & 47.608 & \textbf{0.286} & \textbf{26.057} & \textbf{0.89} \\
    PASD + Ours & \textbf{0.612} & \textbf{0.376} & \textbf{56.463} & 0.371 & 25.159 & 0.87 \\
    \bottomrule
\end{tabular}
    }
    \caption{The performance of PASD and its BIR-Adapter extension on the RealSR dataset.}
    \label{tab:pasd_bir}
\end{table}

\subsection{Ablation studies} \label{sec:ablation}

In \cref{sec:guidance}, we presented a guidance mechanism to mitigate hallucinations. We report results without guidance ($\xi = 1$); its quantitative and qualitative effects are shown in \cref{fig:w_wo_guidance}. We further evaluate two non-fine-tuned variants to assess our design choices at the attention and BIR-Adapter levels. The first replaces self-attention with cross-attention, while the second retains the original attention structure but copies weights into the BIR-Adapter. Both variants fail; thus, we provide results only in the appendix. These experiments confirm the necessity of both the BIR-Adapter formulation and weight fine-tuning.

\section{Results \& discussion}

\subsection{Restoration quality}

\cref{tab:div2k} presents the performance of the baselines and our approach under synthetic degradations. For reference-free metrics, our approach is comparable to the state of the art across all scenarios, both with and without guidance. Notably, in the challenging case of $\downarrow_4 + \sigma_n$, our method achieves state-of-the-art performance. For reference-based metrics, our method remains competitive. This trend also holds for real-world degradations, as shown in \cref{tab:realsr} and \cref{tab:real60}.

A notable observation is the sudden drop in reference-based metrics when guidance is disabled. Although the perceptual quality is higher, as indicated by reference-free metrics, inconsistencies remain. \Cref{fig:w_wo_guidance} displays this for an example image. Conversely, better reference-based performance with worse reference-free performance would not imply better restoration, as confirmed by SwinIR's overly smooth or artifact-prone results. Qualitative results in \Cref{fig:qual} further illustrate how our method produces finer details. Furthermore, we show in \Cref{tab:pasd_bir} the performance of PASD with our approach on the RealSR dataset. The plug-and-play achieves improved image quality and produces sharper restorations, as illustrated by the example output in \Cref{fig:qual}. For more visual results, please refer to the appendix. Some baseline results differ from those originally reported; we provide references to the codebase and evaluation details in the appendix. To compare with the closed-source DR-Adapter~\cite{liang2025diffusion}, we evaluate our model on a real-world dataset and report the results in \cref{tab:real60}.

\subsection{Parameter efficiency \& inference time}

Importantly, as shown in \Cref{tab:parameters}, BIR-Adapter achieves competitive performance while requiring $10\times$ and $36\times$ fewer additional trained parameters than DiffBIR and SUPIR, respectively. While the backbone remains frozen during training, all parameters are used during inference; thus, the reported reduction reflects training rather than the total model size.

BIR-Adapter exhibits inference times comparable to SUPIR despite delivering stronger perceptual quality. A standard SD-1.5 ControlNet-based pipeline requires $486$ GMacs per image, compared to $481$ GMacs of BIR-Adapter, indicating that the runtime overhead stems primarily from the batched processing, rather than architectural complexity. This design enables the reuse of the pretrained diffusion prior without requiring the training of auxiliary feature extractors or full fine-tuning. Degraded features remain static, hence caching can eliminate redundant computation; preliminary experiments have reduced inference time to approximately two seconds under identical conditions.

\section{Conclusion}

We argued that the latent representations of a diffusion prior can be leveraged for restoration. To that end, we introduced a few-parameter restoring attention to a frozen backbone, along with a guidance mechanism. Extensive experiments on synthetic and real-world benchmarks demonstrate that BIR-Adapter achieves competitive, and in several cases superior, restoration quality compared to state-of-the-art, while reducing the number of trained parameters by up to $36 \times$. In addition, the adapted guidance mechanism effectively mitigates hallucinations by balancing perceptual quality and fidelity. The adapter-based design further enables plug-and-play into existing diffusion backbones, as demonstrated by extending a super-resolution–only model to restore unknown degradations without retraining.

Despite these advantages, our approach has limitations. BIR-Adapter significantly reduces the number of trained parameters but introduces overhead because degraded latents are processed in parallel for restoring attention. This might hinder large-scale deployment. Furthermore, the guidance parameter is selected empirically as a hyperparameter. Third, while the method generalizes to real-world degradations, it has not been explicitly evaluated on motion blur or low-light conditions, limiting the scope of experimental validation.

These limitations also suggest promising directions for future research. Since degraded features remain constant across diffusion steps, caching strategies could improve inference efficiency without architectural changes. Moreover, incorporating degradation-adaptive attention to reduce the effective sequence length offers a complementary direction for improving computational efficiency. Automatically adapting the guidance strength based on image content is another important avenue to reduce manual hyperparameter tuning. Finally, extending BIR-Adapter to recent pretrained architectures, such as diffusion transformers, and evaluating it on a broader range of degradations would further strengthen its applicability to real-world blind image restoration tasks.

\section{Declaration of generative AI and AI-assisted technologies in the writing process}

During the preparation of this work the authors used ChatGPT in order to refine sentences and perform a grammar check. After using this tool/service, the authors reviewed and edited the content as needed and take full responsibility for the content of the published article.

\newpage
\appendix
\section{More Visual Results}

\begin{figure}[h!]
    \centering
    \begin{subfigure}[b]{\linewidth}
        \centering
        \includegraphics[width=\linewidth]{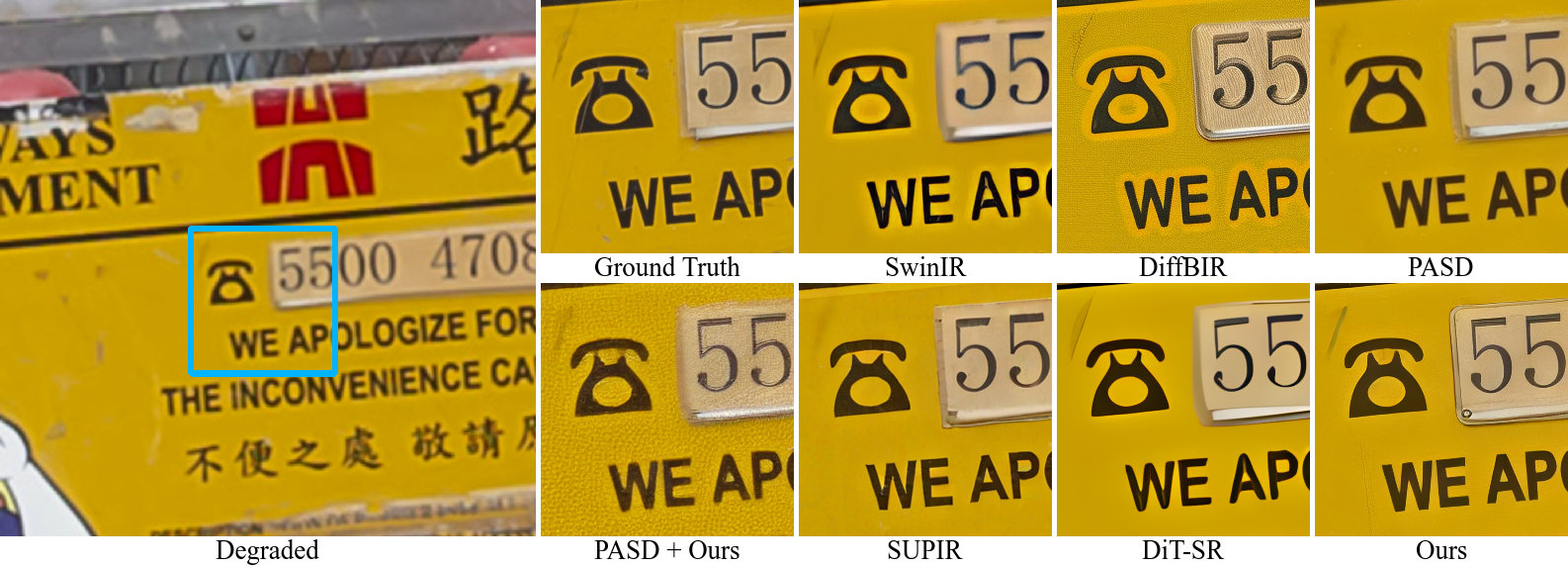}
    \end{subfigure}

    \begin{subfigure}[b]{\linewidth}
        \centering
        \includegraphics[width=\linewidth]{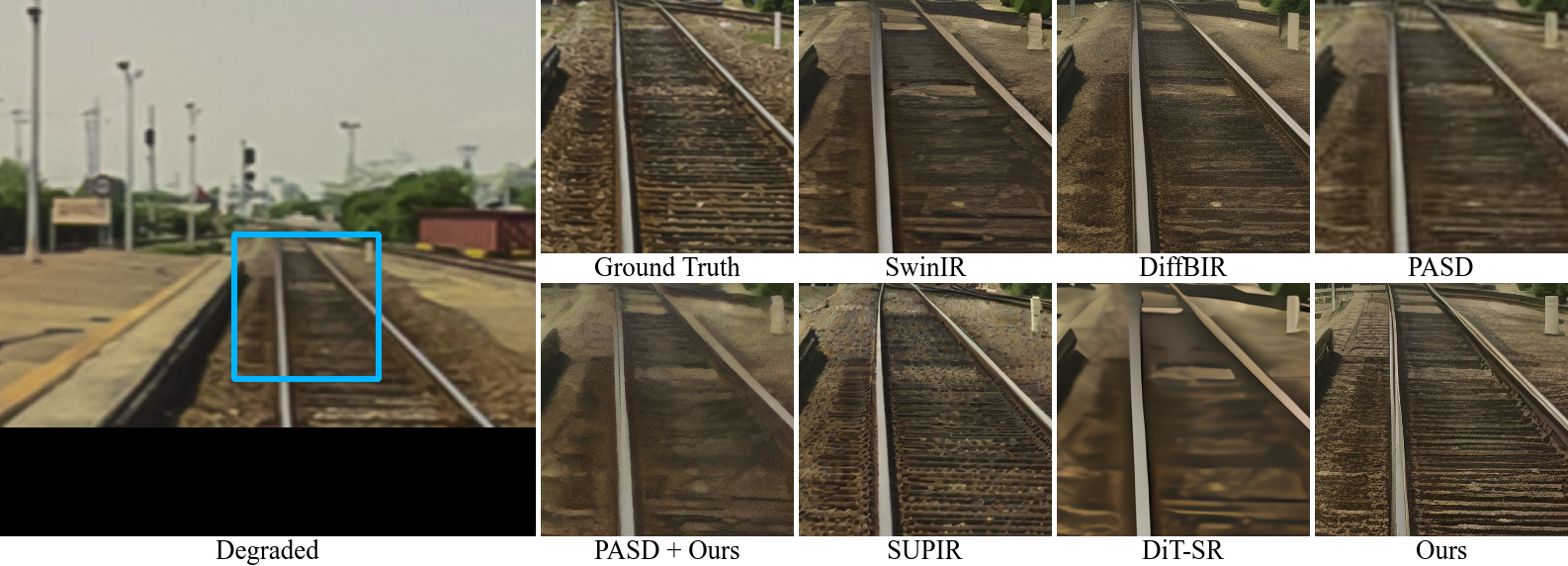}
    \end{subfigure}

    \begin{subfigure}[b]{\linewidth}
        \centering
        \includegraphics[width=\linewidth]{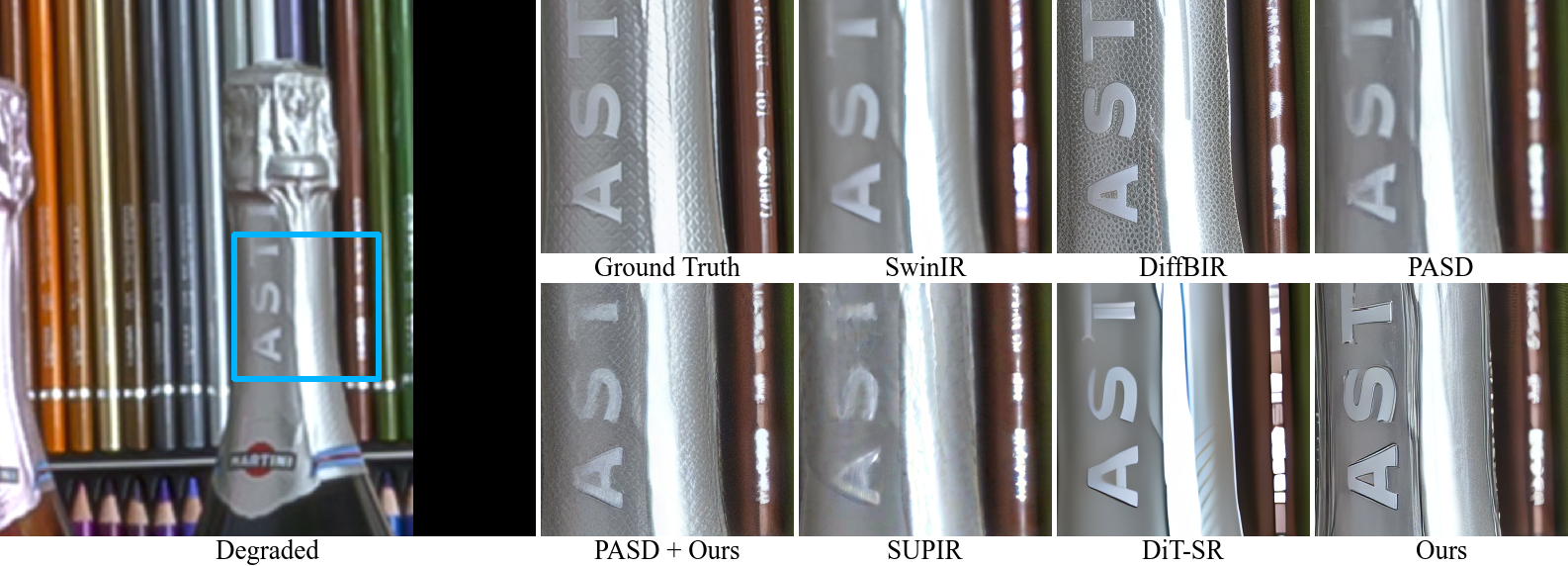}
    \end{subfigure}
    
    \caption{Visual results from RealSR with real-world unknown degradations \cite{cai2019toward}. All images are restored with $4\times$ upsampling using baseline methods and ours. We also include results from the extended PASD.}
\end{figure}
\newpage
\begin{figure}[h!]
    \centering
    \begin{subfigure}[b]{\linewidth}
        \centering
        \includegraphics[width=\linewidth]{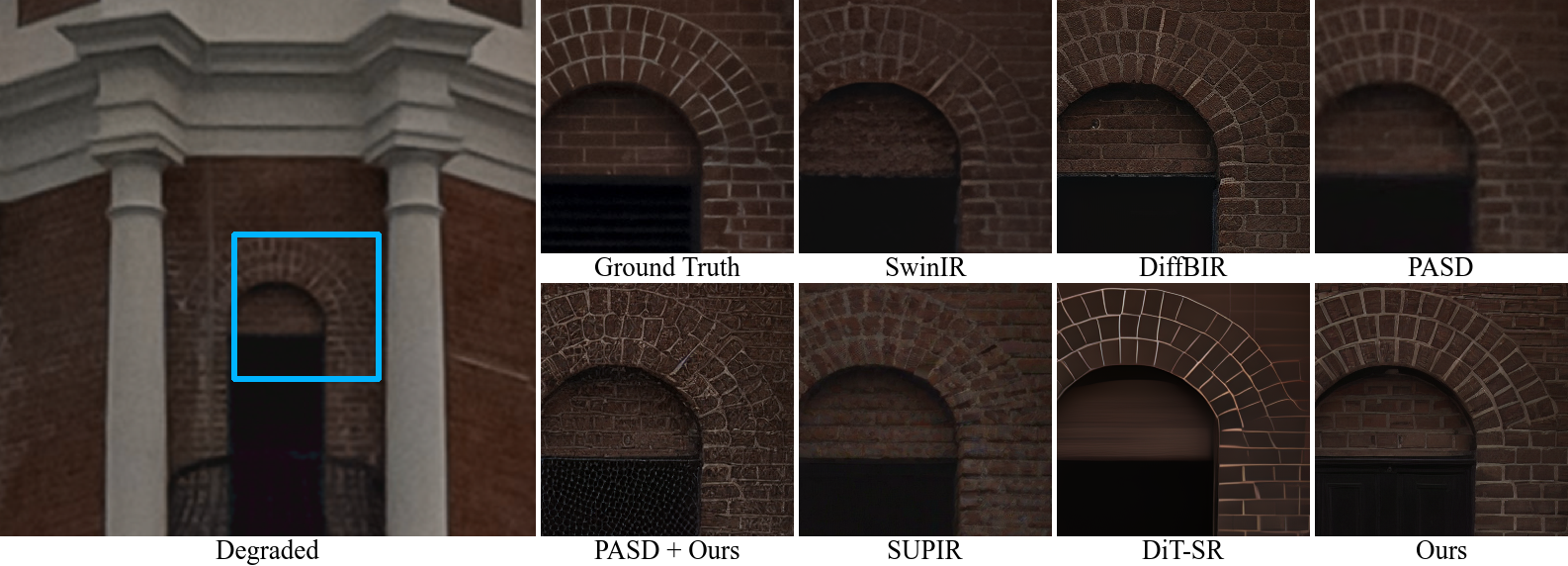}
    \end{subfigure}

    \begin{subfigure}[b]{\linewidth}
        \centering
        \includegraphics[width=\linewidth]{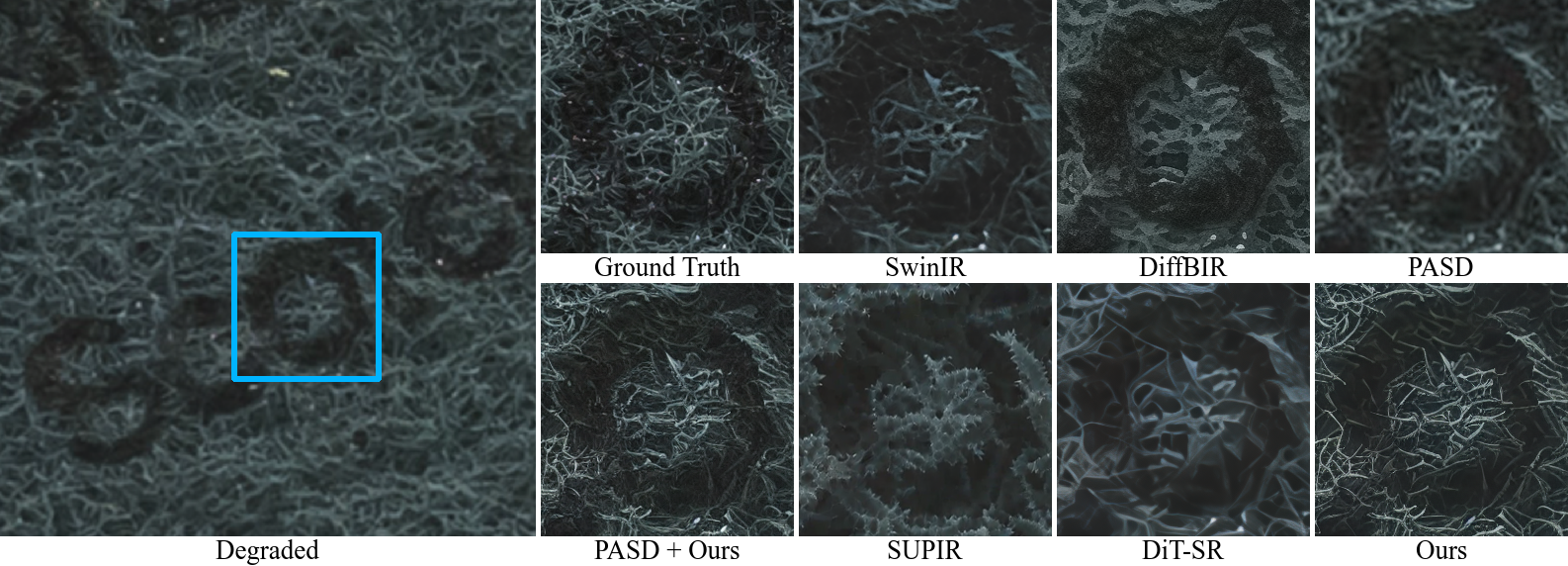}
    \end{subfigure}

    \begin{subfigure}[b]{\linewidth}
        \centering
        \includegraphics[width=\linewidth]{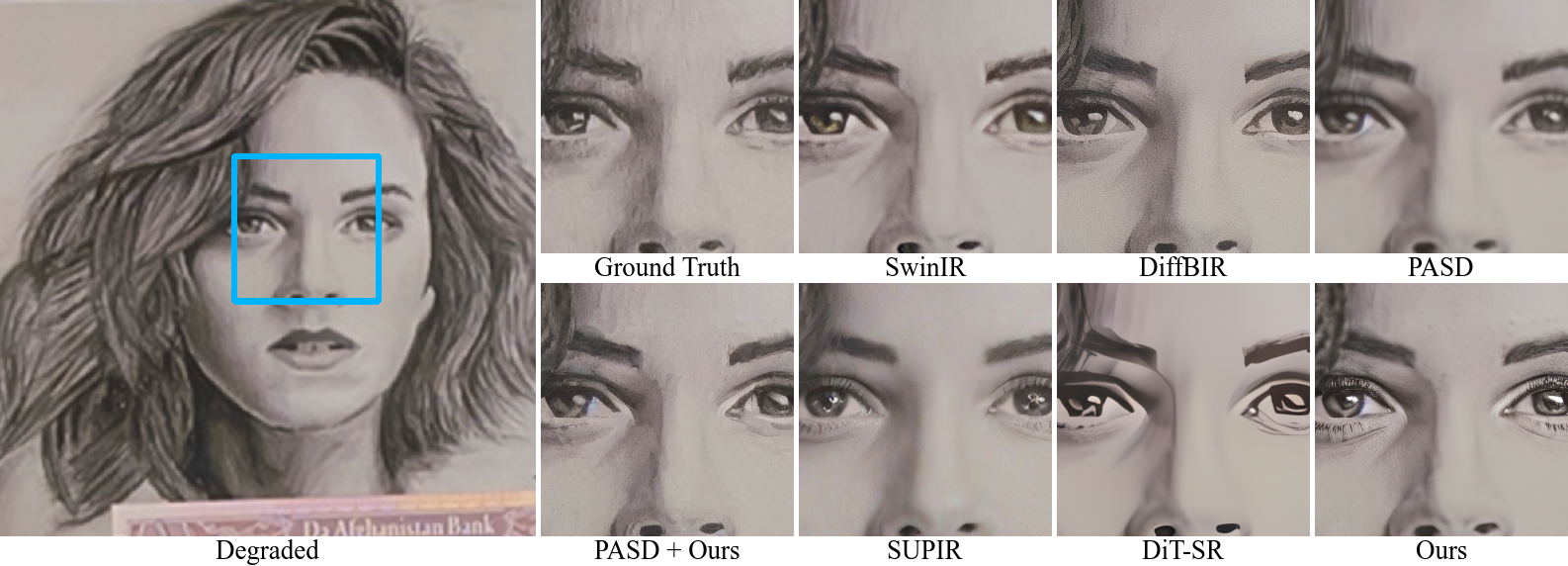}
    \end{subfigure}
    
    \caption{Visual results from RealSR with real-world unknown degradations \cite{cai2019toward}. All images are restored with $4\times$ upsampling using baseline methods and ours. We also include results from the extended PASD.}
\end{figure}
\newpage
\begin{figure}[h!]
    \centering
    \begin{subfigure}[b]{\linewidth}
        \centering
        \includegraphics[width=\linewidth]{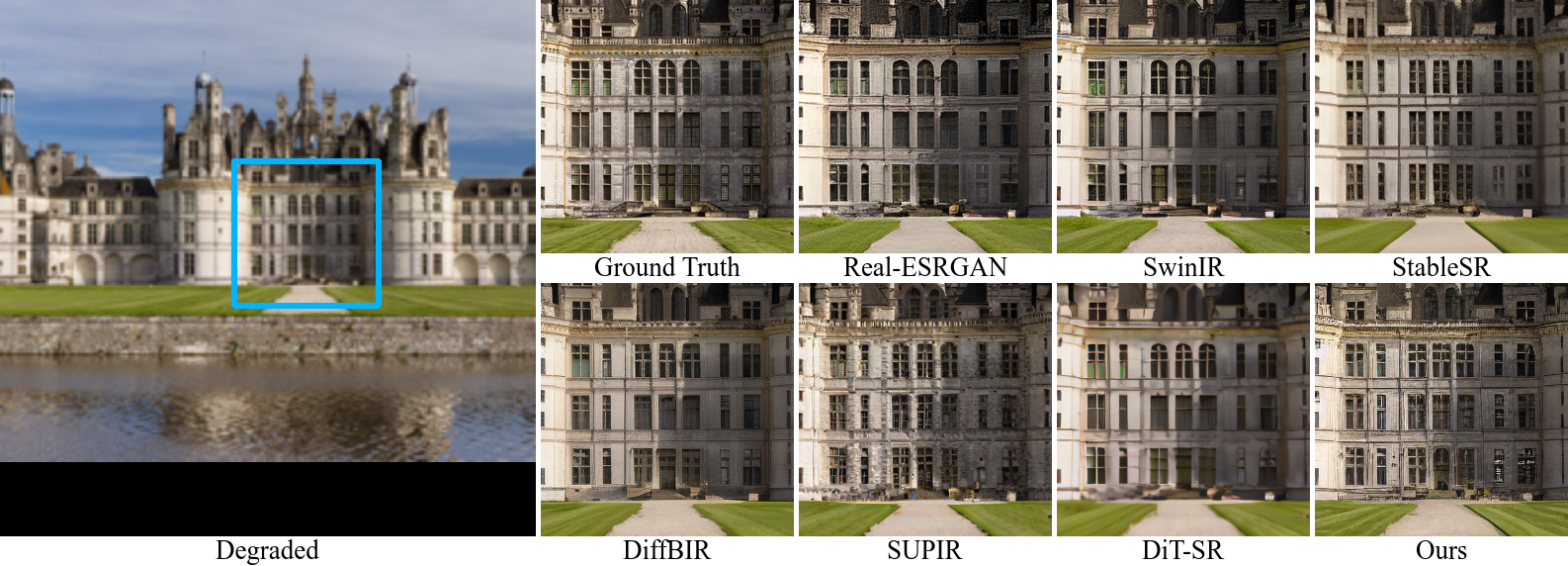}
        \caption{$\sigma_b=2 + \downarrow_4$}
    \end{subfigure}

    \begin{subfigure}[b]{\linewidth}
        \centering
        \includegraphics[width=\linewidth]{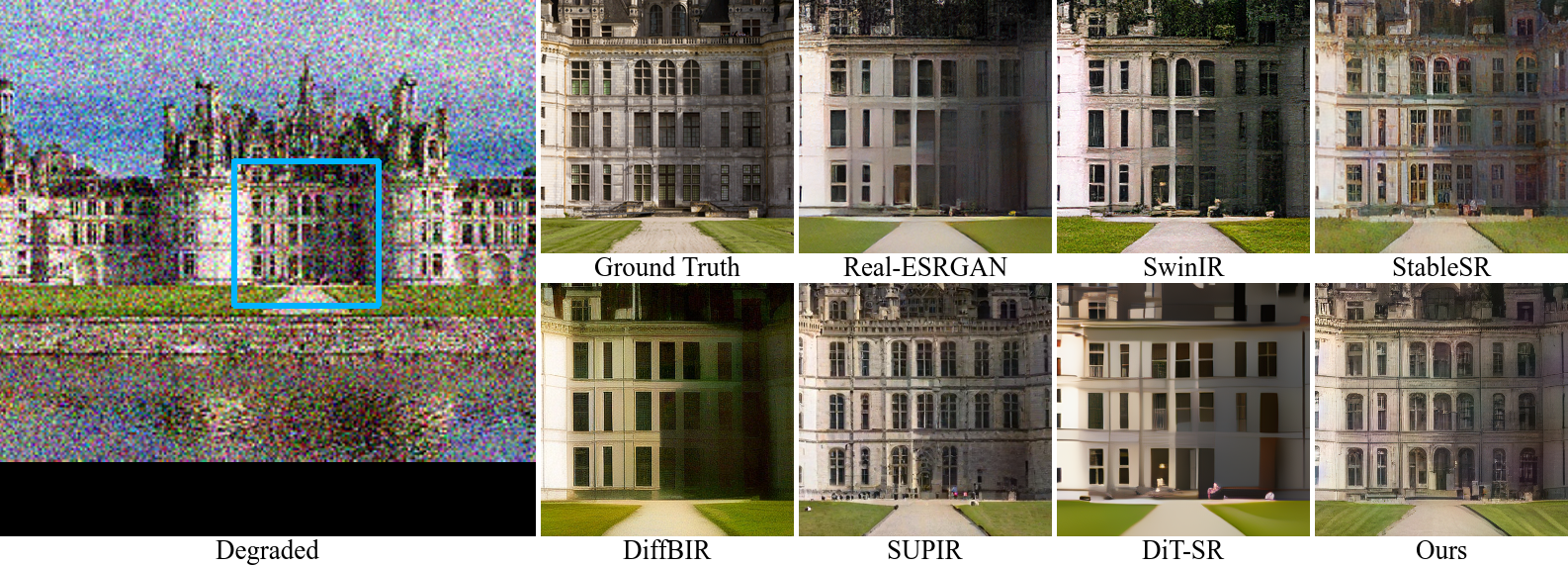}
        \caption{$\downarrow_4 + \sigma_n=40$}
    \end{subfigure}

    \begin{subfigure}[b]{\linewidth}
        \centering
        \includegraphics[width=\linewidth]{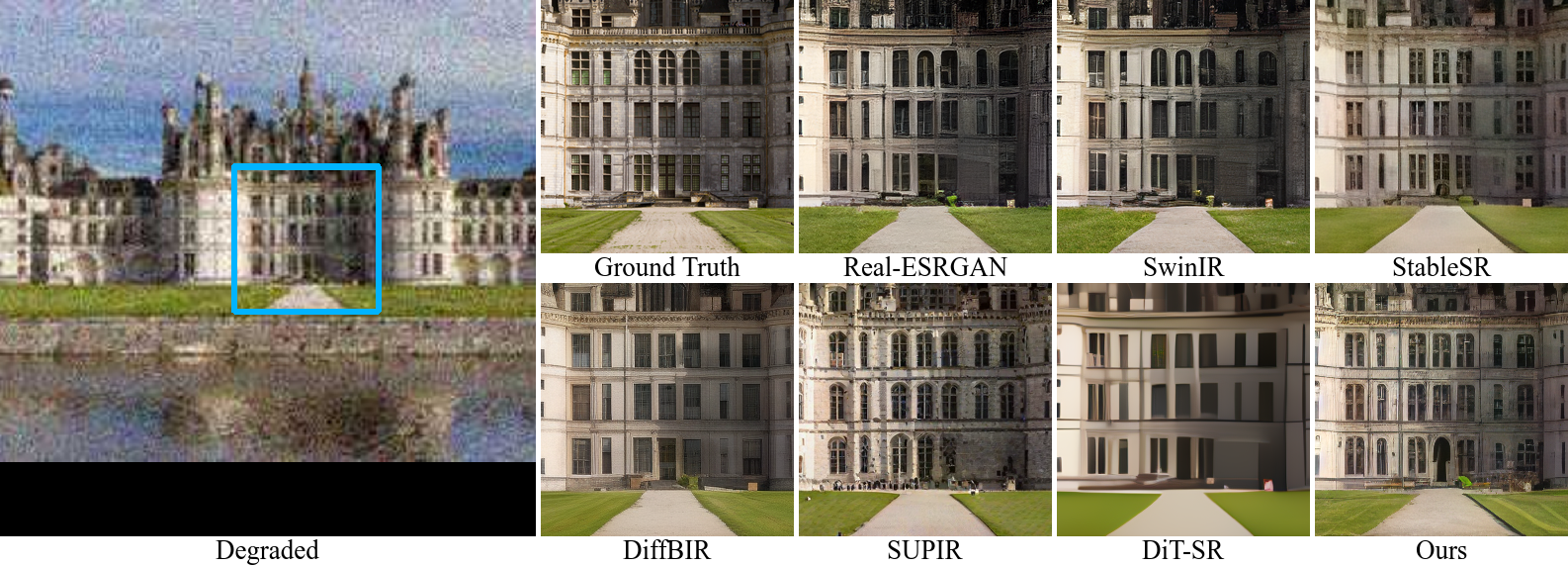}
        \caption{$\sigma_b=2 + \downarrow_4 + \sigma_n=20 + Q=50$}
    \end{subfigure}
    
    \caption{Visual results of baseline methods and ours on a sample frame from the DIV2K validation set under synthetic degradations \cite{agustsson2017ntire}.}
\end{figure}
\newpage
\begin{figure}[h!]
    \centering
    \begin{subfigure}[b]{\linewidth}
        \centering
        \includegraphics[width=\linewidth]{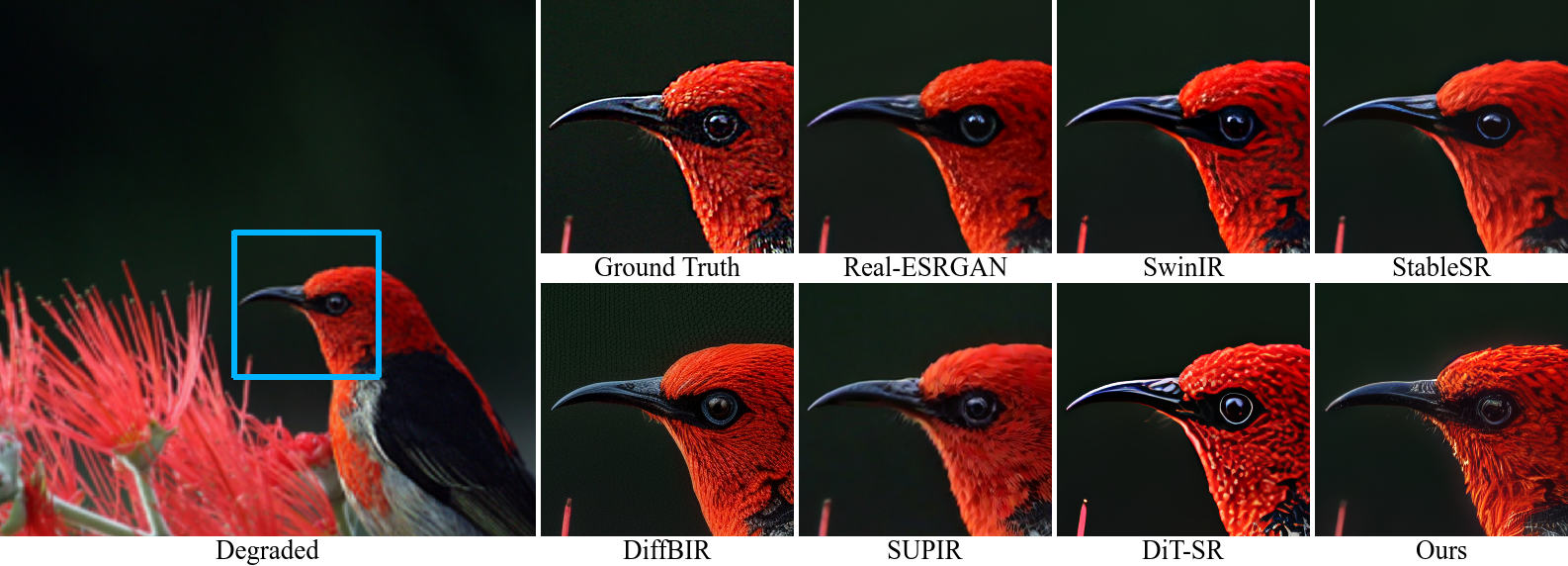}
        \caption{$\sigma_b=2 + \downarrow_4$}
    \end{subfigure}

    \begin{subfigure}[b]{\linewidth}
        \centering
        \includegraphics[width=\linewidth]{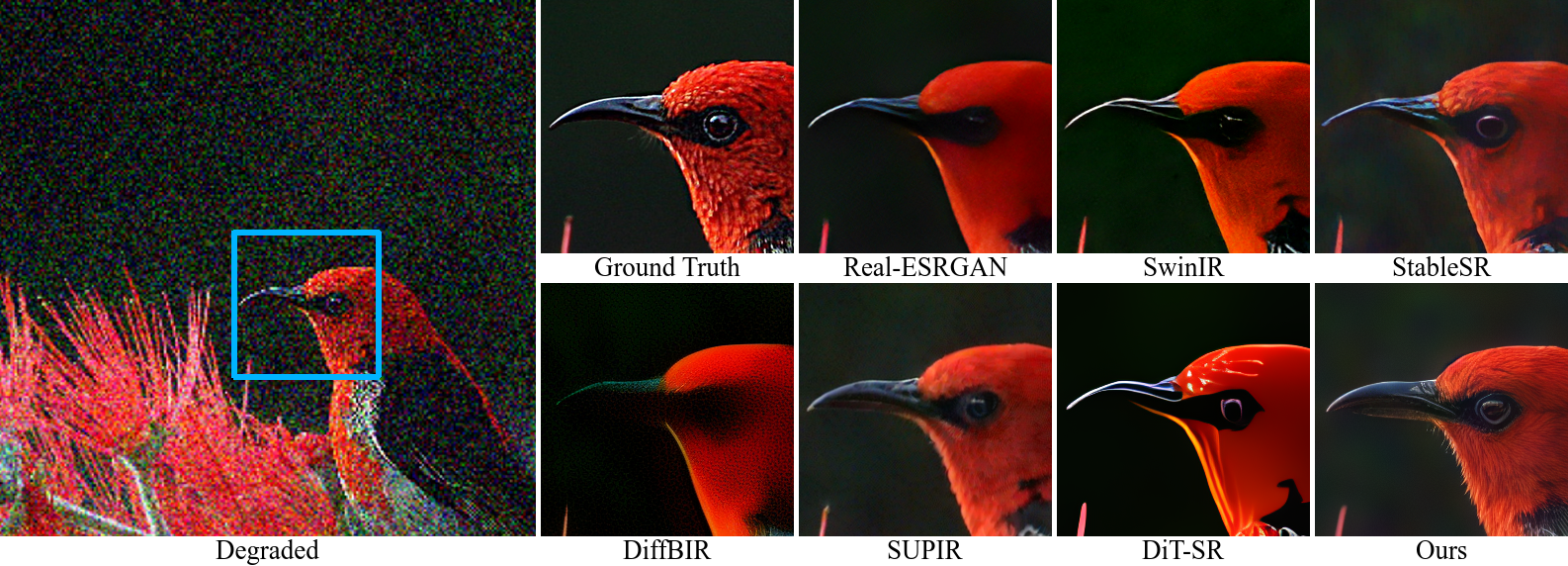}
        \caption{$\downarrow_4 + \sigma_n=40$}
    \end{subfigure}

    \begin{subfigure}[b]{\linewidth}
        \centering
        \includegraphics[width=\linewidth]{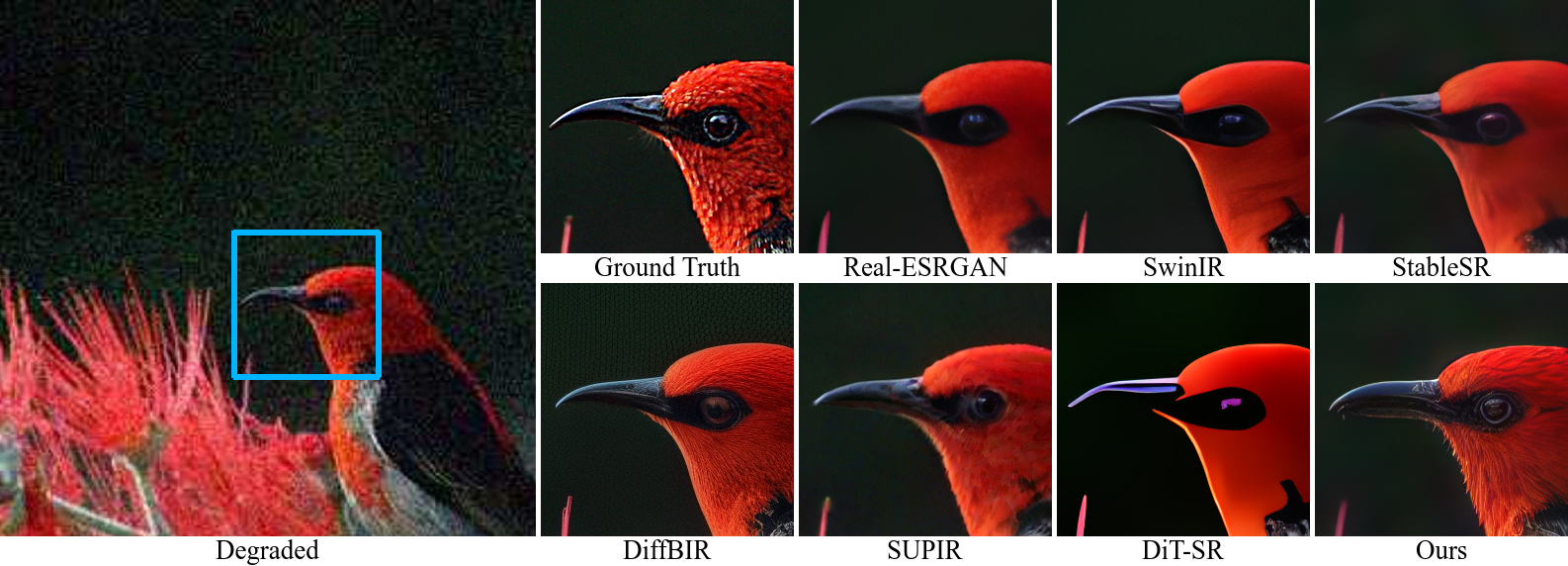}
        \caption{$\sigma_b=2 + \downarrow_4 + \sigma_n=20 + Q=50$}
    \end{subfigure}
    
    \caption{Visual results of baseline methods and ours on a sample frame from the DIV2K validation set under synthetic degradations \cite{agustsson2017ntire}.}
\end{figure}
\newpage
\begin{figure}[h!]
    \centering
    \begin{subfigure}[b]{\linewidth}
        \centering
        \includegraphics[width=\linewidth]{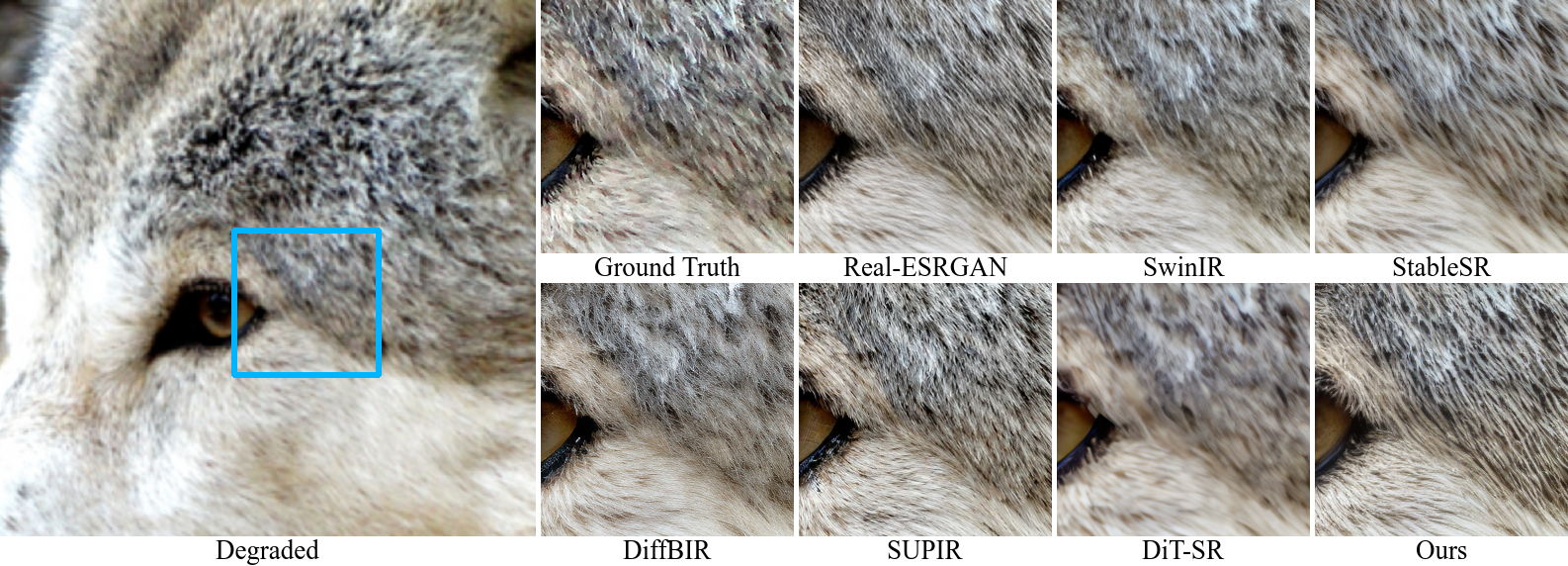}
        \caption{$\sigma_b=2 + \downarrow_4$}
    \end{subfigure}

    \begin{subfigure}[b]{\linewidth}
        \centering
        \includegraphics[width=\linewidth]{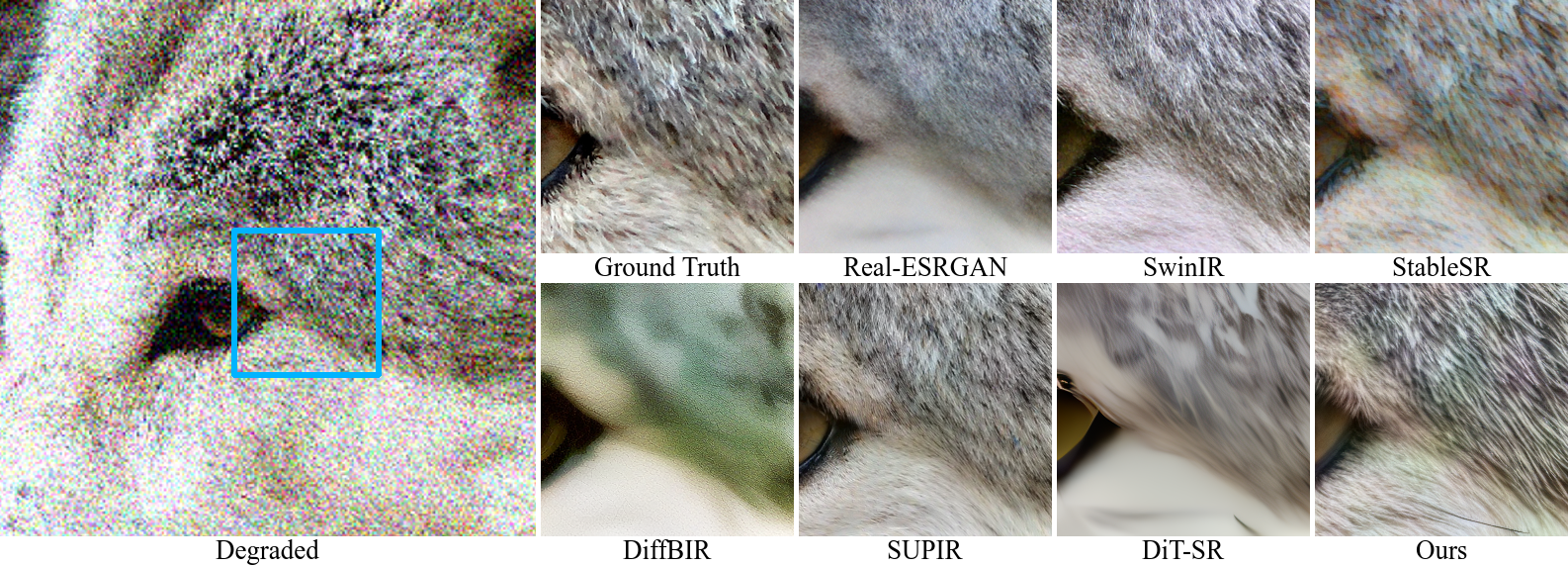}
        \caption{$\downarrow_4 + \sigma_n=40$}
    \end{subfigure}

    \begin{subfigure}[b]{\linewidth}
        \centering
        \includegraphics[width=\linewidth]{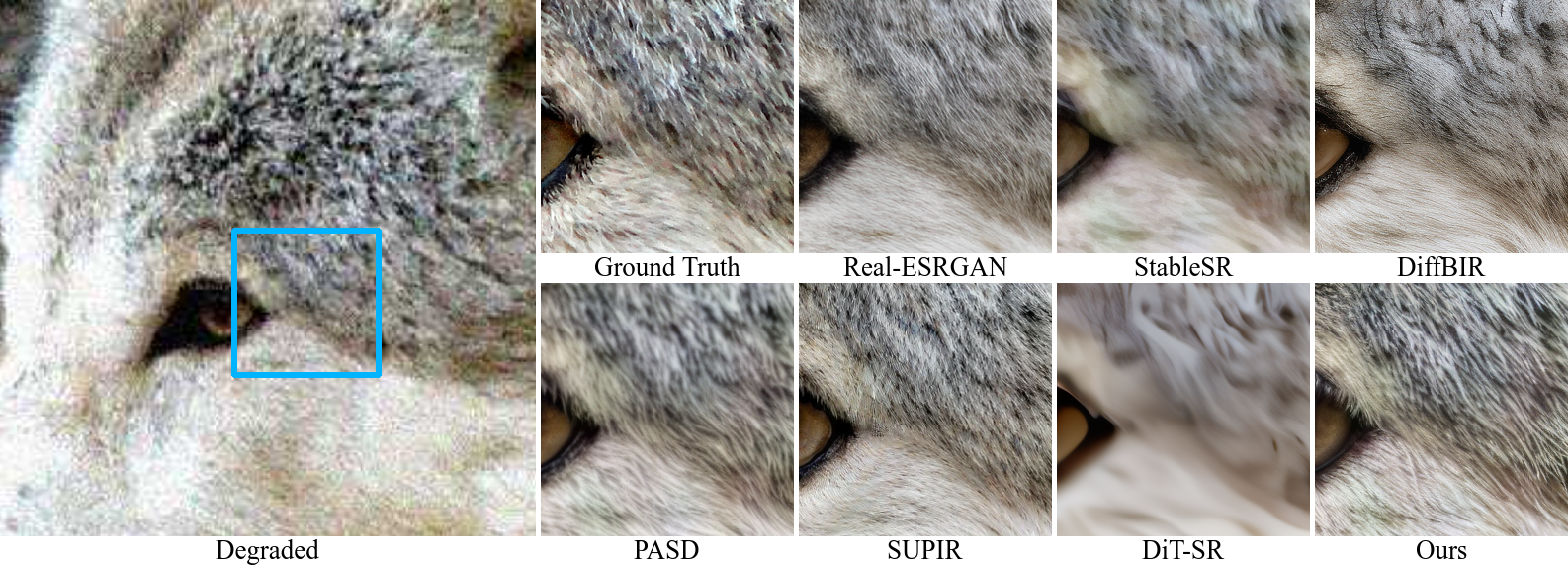}
        \caption{$\sigma_b=2 + \downarrow_4 + \sigma_n=20 + Q=50$}
    \end{subfigure}
    
    \caption{Visual results of baseline methods and ours on a sample frame from the DIV2K validation set under synthetic degradations \cite{agustsson2017ntire}.}
\end{figure}

\section{Training-free Variants and Pseudocodes}

\begin{figure}[H]
    \centering
    \begin{subfigure}[b]{0.24\linewidth}
        \centering
        \includegraphics[width=\linewidth]{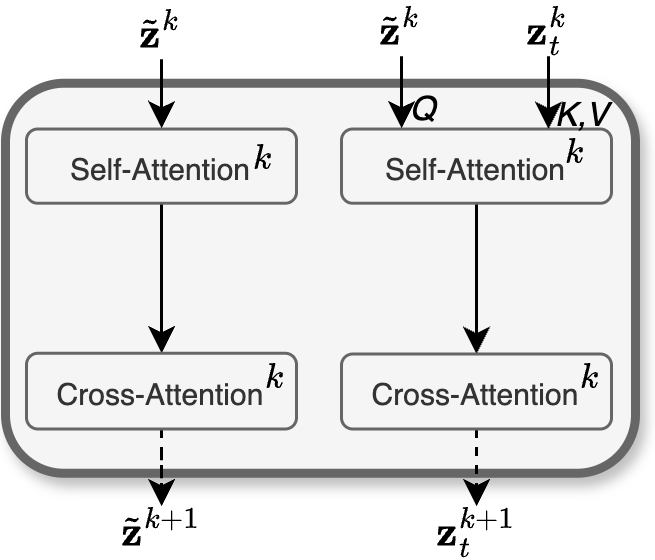}
        \caption{Variant 1}
        \label{fig:var1}
    \end{subfigure}
    \begin{subfigure}[b]{0.24\linewidth}
        \centering
        \includegraphics[width=\linewidth]{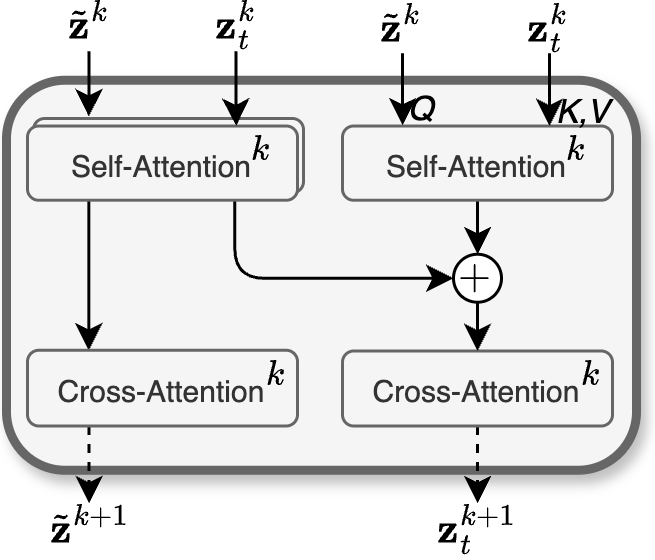}
        \caption{Variant 2}
        \label{fig:var2}
    \end{subfigure}
    \begin{subfigure}[b]{0.5\linewidth}
        \centering
        \includegraphics[width=\linewidth]{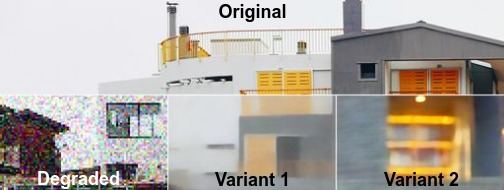}
        \caption{Restoration using variants}
        \label{fig:var_res}
    \end{subfigure}

    \caption{Variant 1 utilizes the degraded features in the self-attention layer as query. Variant 2 does not fine-tune the weights of BIR-Adapter. Restoration results in \subref{fig:var_res} show how the variants fail.}
\label{fig:vars}
\end{figure}

\begin{algorithm}
{\tiny
\caption{Pseudocode of BIR-Adapter and its variants.}
\begin{algorithmic}
\STATE \textbf{procedure} \textsc{$\epsilon_\theta$}($\tilde{\mathbf{x}}, \mathbf{x}_t, y^+$)
\STATE $\mathbf{X} \leftarrow \text{concat}[\tilde{\mathbf{x}}, \mathbf{x}_t]$

\STATE \textbf{BIR-Adapter}
\FOR{$k \gets 1$ to last layer $K$}    
    \STATE Extract hidden state $\mathbf{Z}^k$ of layer $k$ from $\mathbf{Z}^{k-1}$, where $\mathbf{Z}^{0}$ is $\mathbf{X}$ // Through convolution, pooling, etc.
    \STATE $\tilde{\mathbf{z}}^k$, $\mathbf{z}^k_t \leftarrow \mathrm{chunk}(\mathbf{Z}^k, 2)$    
    \STATE $\tilde{Q} \gets \tilde{\mathbf{z}}^k \mathbf{W}^Q, \quad K' \gets \mathbf{z}^k_t \mathbf{W'}^K, \quad V' \gets \mathbf{z}^k_t \mathbf{W'}^V, \quad Q \gets \mathbf{Z}^k\mathbf{W}^Q, \quad K \gets \mathbf{Z}^k\mathbf{W}^K, \quad V \gets \mathbf{Z}^k\mathbf{W}^V$
    \STATE $\mathbf{Z}^{k+1} \gets Attention(Q, K, V)\mathbf{W}^O + Attention(\tilde{Q}, K', V')\mathbf{W'}^O$
    \STATE Execute cross-attention between $\mathbf{Z}^{k+1}$ and $y^+$.
\ENDFOR

\vspace{0.5em} 

\STATE \textbf{Variant 1: Degraded features as the query at Self-Attention}

\FOR{$k \gets 1$ to last layer $K$}
    \STATE Extract hidden state $\mathbf{Z}^k$ of layer $k$ from $\mathbf{Z}^{k-1}$, where $\mathbf{Z}^{0}$ is $\mathbf{X}$ // Through convolution, pooling, etc.
    \STATE $\tilde{Q} \gets \tilde{\mathbf{z}}^k \mathbf{W}^Q, \quad Q \gets \mathbf{Z}^k\mathbf{W}^Q, \quad K \gets \mathbf{Z}^k\mathbf{W}^K, \quad V \gets \mathbf{Z}^k\mathbf{W}^V$
    \STATE $\mathbf{Z}^{k+1} \gets Attention(\tilde{Q}, K, V)\mathbf{W}^O$
    \STATE Execute cross-attention between $\mathbf{Z}^{k+1}$ and $y^+$.
\ENDFOR

\vspace{0.5em} 

\STATE \textbf{Variant 2: Non-trained BIR-Adapter}
\FOR{$k \gets 1$ to last layer $K$}
    \STATE Extract hidden state $\mathbf{Z}^k$ of layer $k$ from $\mathbf{Z}^{k-1}$, where $\mathbf{Z}^{0}$ is $\mathbf{X}$ // Through convolution, pooling, etc.
    \STATE $\tilde{Q} \gets \tilde{\mathbf{z}}^k \mathbf{W}^Q, \quad Q \gets \mathbf{Z}^k\mathbf{W}^Q, \quad K \gets \mathbf{Z}^k\mathbf{W}^K, \quad V \gets \mathbf{Z}^k\mathbf{W}^V$
    \STATE $\mathbf{Z}^{k+1} \gets Attention(Q, K, V)\mathbf{W}^O + Attention(\tilde{Q}, K, V)\mathbf{W}^O$
    \STATE Execute cross-attention between $\mathbf{Z}^{k+1}$ and $y^+$.
\ENDFOR
\STATE \textbf{return} final prediction $\mathbf{z}_t^{K}$
\STATE \textbf{end procedure}
\end{algorithmic}
}
\end{algorithm}
\newpage
\section{Baselines}

\begin{figure}[h!]
\centering
\begin{minipage}{1\textwidth}
\begin{lstlisting}[style=shell]
export DEFAULT_PROMPT = "Cinematic, High Contrast, highly detailed, taken using a Canon EOS R camera, hyper detailed photo - realistic maximum detail, 32k, Color Grading, ultra HD, extreme meticulous detailing, skin pore detailing, hyper sharpness, perfect without deformations."
export NEGATIVE_PROMPT = "painting, oil painting, illustration, drawing, art, sketch, oil painting, cartoon, CG Style, 3D render, unreal engine, blurring, dirty, messy, worst quality, low quality, frames, watermark, signature, jpeg artifacts, deformed, lowres, over-smooth"
\end{lstlisting}
\begin{lstlisting}[style=shell]
git clone https://github.com/XPixelGroup/DiffBIR.git

python inference.py --task sr --version v2 --upscale 4 --sampler spaced --cfg_scale 4.0 --steps 20 
    --captioner none --input /input --output /output --precision fp16 --cleaner_tiled 
    --vae_encoder_tiled --vae_decoder_tiled --cldm_tiled --seed 1234 
    --pos_prompt DEFAULT_PROMPT --neg_prompt NEGATIVE_PROMPT

\end{lstlisting}
\begin{lstlisting}[style=shell]
git clone https://github.com/yangxy/PASD.git

python test_pasd.py --pretrained_model_path /path/to/sd --pasd_model_path /pasd/model 
    --num_inference_steps 20 --upscale 4 --high_level_info none --process_size 512 
    --image_path /input --output_dir /output --added_prompt DEFAULT_PROMPT 
    --negative_prompt NEGATIVE_PROMPT --seed 1234

\end{lstlisting}
\begin{lstlisting}[style=shell]
git clone hhttps://github.com/Fanghua-Yu/SUPIR.git

python test.py --edm_steps 20 --upscale 4 --no_llava --s_stage2 0.93 --s_cfg 6.0 --spt_linear_CFG 3.0 
    --s_noise 1.02 --img_dir /input --save_dir /output --ae_dtype bf16 --diff_dtype bf16 
    --use_tile_vae --encoder_tile_size 1024 --decoder_tile_size 128 --a_prompt DEFAULT_PROMPT 
    --n_prompt NEGATIVE_PROMPT 

\end{lstlisting}
\begin{lstlisting}[style=shell]
git clone https://github.com/IceClear/StableSR.git

from predict import Predictor
predictor = Predictor()
for degraded_image in /path/to/degraded/images:
    out_file = predictor.predict(input_image=degraded_image, ddpm_steps=20, fidelity_weight=0.5, 
                                 upscale=4, tile_overlap=32, colorfix_type='wavelet', seed=1234)

\end{lstlisting}
\begin{lstlisting}[style=shell]
git clone https://github.com/kunncheng/DiT-SR.git

python inference.py --task realsr --scale 4 --config_path configs/realsr_DiT.yaml 
    --ckpt_path /path/to/DiT-SR --chop_size 512 --chop_stride 256 -i /input -o /output 

\end{lstlisting}
\captionof{lstlisting}{Source codes and the commands used to create the results of the baselines.}
\end{minipage}
\end{figure}
\newpage

\bibliographystyle{elsarticle-num} 
\bibliography{main.bib}

\end{document}